\documentclass[aps, prx, twocolumn, floatfix, superscriptaddress, reprint, 10pt]{revtex4-2}

\usepackage{graphicx}\usepackage{verbatim}
\usepackage[version=4]{mhchem}
\usepackage{xcolor}
\usepackage{physics}
\usepackage[normalem]{ulem}
\usepackage{pdfpages}
\usepackage{pgffor}
\usepackage{soul}
\usepackage{hyperref}
\usepackage{amsmath,amssymb}
\usepackage{booktabs}

\newif\ifhighlight

\newcommand{\highlight}{\highlighttrue}
\highlight

\newcommand{\editor}[2]{\expandafter\newcommand\csname #1note\endcsname[1]{\textcolor{#2}{(\textbf{#1note:} \textsc{##1})}}\expandafter\newcommand\csname #1\endcsname[1]{\ifhighlight\textcolor{#2}{##1} \else ##1\fi}\expandafter\newcommand\csname #1cancel\endcsname[1]{\ifhighlight\textcolor{#2}{\sout{##1}}\fi}\expandafter\newcommand\csname #1change\endcsname[2]{\ifhighlight\textcolor{#2}{\sout{##1} ##2}\else ##2\fi}\newenvironment{#1text}{\ifhighlight\color{#2}\fi}{\color{black}}
}
\definecolor{tangerine}{rgb}{0.944,0.522,0}
\definecolor{verde}{rgb}{0.,0.6,0}

\editor{resub}{cyan}
\editor{FB}{teal}
\editor{MC}{red}
\editor{DT}{verde}
\editor{PP}{tangerine}
\editor{JA}{blue}
\editor{MD}{purple}

\makeatletter
\AtBeginDocument{\let\LS@rot\@undefined}
\makeatother

\begin{document}

\title{How unconstrained machine-learning models learn physical symmetries}

\author{M. Domina}
\thanks{M. Domina, J. W. Abbott and P. Pegolo contributed equally to this work.}
\affiliation{Laboratory of Computational Science and Modeling, Institut des Mat\'eriaux, \'Ecole Polytechnique F\'ed\'erale de Lausanne, 1015 Lausanne, Switzerland}

\author{J. W. Abbott}
\thanks{M. Domina, J. W. Abbott and P. Pegolo contributed equally to this work.}
\affiliation{Laboratory of Computational Science and Modeling, Institut des Mat\'eriaux, \'Ecole Polytechnique F\'ed\'erale de Lausanne, 1015 Lausanne, Switzerland}

\author{P. Pegolo}
\thanks{M. Domina, J. W. Abbott and P. Pegolo contributed equally to this work.}
\affiliation{Laboratory of Computational Science and Modeling, Institut des Mat\'eriaux, \'Ecole Polytechnique F\'ed\'erale de Lausanne, 1015 Lausanne, Switzerland}

\author{F. Bigi}
\affiliation{Laboratory of Computational Science and Modeling, Institut des Mat\'eriaux, \'Ecole Polytechnique F\'ed\'erale de Lausanne, 1015 Lausanne, Switzerland}

\author{M. Ceriotti}
\affiliation{Laboratory of Computational Science and Modeling, Institut des Mat\'eriaux, \'Ecole Polytechnique F\'ed\'erale de Lausanne, 1015 Lausanne, Switzerland}

\newcommand{\mc}[1]{{\color{blue}#1}}

\date{\today}

\begin{abstract}
The requirement of generating predictions that exactly fulfill the fundamental symmetry of the corresponding physical quantities has profoundly shaped the development of machine-learning models for physical simulations.
In many cases, models are built using constrained mathematical forms that ensure that symmetries are enforced exactly. 
However, unconstrained models that do not obey rotational symmetries are often found to have competitive performance, and to be able to \emph{learn} to a high level of accuracy an approximate equivariant behavior with a simple data augmentation strategy.
In this paper, we introduce rigorous metrics to measure the symmetry content of the learned representations in such models, and assess the accuracy by which the outputs fulfill the equivariant condition.
We apply these metrics to two unconstrained, transformer-based models operating on decorated point clouds (a graph neural network for atomistic simulations and a PointNet-style architecture for particle physics) to investigate how symmetry information is processed across architectural layers and is learned during training.
Based on these insights, we establish a rigorous framework for diagnosing spectral failure modes in ML models. Enabled by this analysis, we demonstrate that one can achieve superior stability and accuracy by strategically injecting the minimum required inductive biases, preserving the high expressivity and scalability of unconstrained architectures while guaranteeing physical fidelity. \end{abstract}

\maketitle
\section{Introduction}

Symmetries are a cornerstone of modern physics. Their profound connection with conservation laws is enshrined in Noether's theorem, a principle that has long guided the formalization of empirical phenomena into theoretical frameworks and served as a foundation for the development of new theories~\cite{Noether1918,Gross1996}. 
Different fields of physical science are characterized by distinct symmetry groups, such as the orthogonal group $\mathrm{O(3)}$ in molecular mechanics~\cite{bart+10prl}, the Lorentz group $\mathrm{SO(1, 3)}$ in high-energy physics~\cite{pelican,lorentzml}, and the special unitary groups $\mathrm{SU}(2^N)$ and $\mathrm{SU}(3)$ in quantum mechanics and chromodynamics, respectively~\cite{latticegaugeml,Aarts2025,batatia2023neurips}.
Consequently, in the development of data-driven models for physics, incorporating physical symmetries has been often regarded as the most natural choice. This approach has led to the rapid and fruitful development of machine-learning (ML) approaches across the physical sciences, with atomistic simulations standing out as a very successful application domain~\cite{behl-parr07prl, bart+10prl,shap16mms,drau19prb,bata+22nips,pozd-ceri23nips}. However, ensuring that a model strictly preserves group equivariance, guaranteeing that outputs transform predictably under group actions on the input, imposes rigid architectural constraints that can be computationally expensive, and limits the expressivity of models~\cite{pozd+20prl,pozd+22mlst}.

Conversely, mainstream computer science and machine learning methods have evolved to maximize architectural expressivity and efficiency, letting domain-specific inductive biases be learned directly from the data. 
Recently, there has been growing interest in applying these ``unconstrained'' models to physics and chemistry, with notable examples being the release of AlphaFold 3, which relaxed the strict equivariant constraints of previous versions~\cite{abra+24nature}, the classification of plasma crystals~\cite{Dormagen2025}, galaxy morphology~\cite{anagnostidis2022cosmologygalaxyredshiftsurveys}, and particle traces~\cite{Young_2026} using PointNet-like architectures~\cite{qi+17ieee}. 
By relaxing strict equivariance constraints, these models aim for increased fitting power and computational efficiency at the expense of having to learn fundamental symmetries from data, typically through data augmentation over symmetry groups~\cite{JMLR:v21:20-163}, with recent works~\cite{lang+24mlst,mazi+25ncomm} showing that errors due to approximate symmetry are negligible compared to the baseline model accuracy.

In the field of atomistic simulations, unconstrained models are gaining traction as successful alternatives to explicitly equivariant models, providing fast, accurate, and transferrable surrogate models for quantum mechanics. 
Machine-learning interatomic potentials (MLIPs), that predict the potential energy surface (PES) (i.e. energies, atomic forces, and cell stresses) from atomic positions and chemical types, currently are, in particular, one of the most active areas in this regard.
Recent works~\cite{pozd-ceri23nips,mazitov_pet-mad_2025,rhodes2025orbv3atomisticsimulationscale,kreiman2025transformersdiscovermolecularstructure} demonstrate that unconstrained models can match or outperform invariant and equivariant architectures on benchmarks, while offering superior scalability through operations that can be implemented efficiently in current accelerated architectures. 
The capability of unconstrained MLIPs extends beyond benchmarks, excelling in realistic scenarios such as complex materials science simulations~\cite{mazitov_pet-mad_2025} and high-throughput screening~\cite{riebesell2025matbenchdiscovery,bigi2026+arxv}.

\begin{figure}[tb]
    \centering
    \centering
    \includegraphics[width=\linewidth]{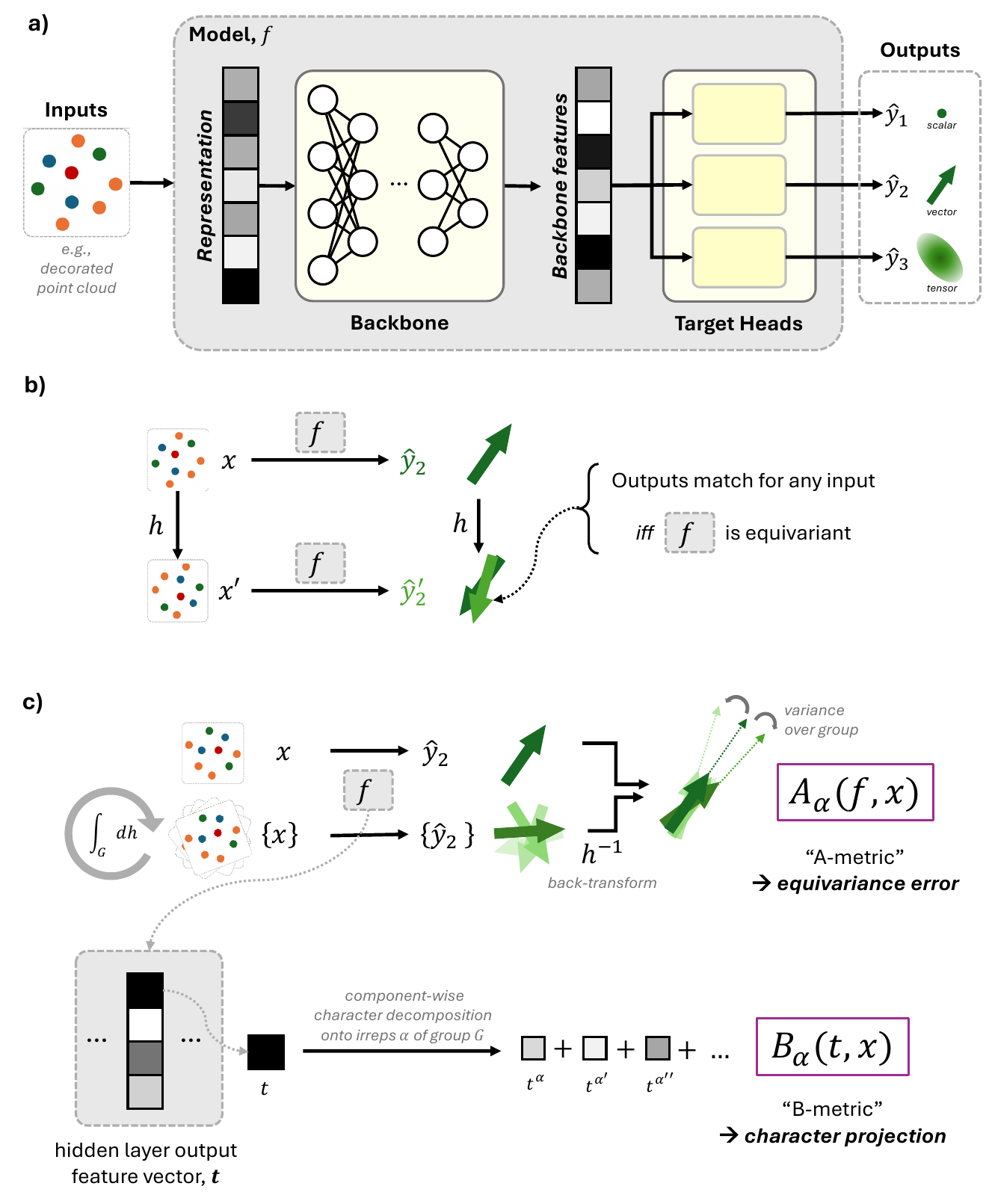}
    \caption{
Overview of the structure of a symmetry-aware ML model, the conditions of group equivariance, and the symmetry diagnostic metrics introduced in this work. a) the ML model is represented by a generic smooth function, $f$, that predicts the physical properties (tensors of different rank), $\hat{y}$, of an input, e.g. a decorated point cloud, $x$. $f$ can be a symmetry-preserving (i.e. equivariant) or unconstrained model. b) Group equivariance is preserved if and only if the model predictions transform like the inputs under the action of the appropriate group symmetry operations. c) The metrics $A_{\alpha}$ and $B_{\alpha}$ introduced in this work quantify the equivariance error of model predictions and the group symmetry content of internal features. For a set of inputs given by Haar integration of $x$ over the group, the \emph{equivariance error}, $A_{\alpha}$, is given by the variance of back-transformed model predictions, while the \emph{character projections}, $B_{\alpha}$, gives the group symmetry decomposition of model features from arbitrary layers.
    }
    \label{fig:fig1}
\end{figure}

More recently, unconstrained architectures have been applied to tasks beyond invariant prediction (e.g., energy), extending to vectorial (e.g., forces~\cite{pmlr-v267-bigi25a,orb}) and higher-rank tensorial targets~\cite{pmlr-v202-duval23a, kell+25jpcl,hua2026scalabledielectrictensorpredictions}. These tasks are significantly more demanding, requiring the model to learn complex transformation laws \textit{via} data augmentation. Unconstrained architectures appear capable of rising to these challenges, though potentially at the cost of longer trainings required to sample the group orbits that equivariant models encode by design~\cite{bigi2026+arxv}.

Given the empirical evidence that these models \emph{can} learn symmetry, natural questions arise: \emph{when} and \emph{how} are these symmetries learned, both across the model architecture and throughout the training process? And can this knowledge drive the design of better performing architectures? 
We introduce metrics to quantify how much the outputs of a model violate the equivariant conditions, as well as to partition the information encoded in the model in terms of the irreducible representations of the relevant groups, and use them to analyze quantitatively the flow of symmetry information across the architecture, and its evolution during training. 
We will then show that, indeed, looking inside the model ``black box'' and understanding the interplay of symmetry and data at every level of the architecture allows one to make informed decisions regarding which inductive biases are essential and which are superfluous, ultimately improving model performance. 

\section{Quantifying symmetry in ML models}

We focus on a class of ML models that operate on decorated point clouds to predict physically relevant quantities (see \autoref{fig:fig1}a), even though similar apply to more general settings, such as 3D shape recognition. Common examples range from invariant quantities, such as potential energy in atomistic simulations, local pressure in fluid dynamics~\cite{lino2022}, or semantic class labels in vision models~\cite{thomas2018tensorfieldnetworksrotation}, to geometric tensors such as fluid velocity fields~\cite{lino2022} and surface normals~\cite{NEURIPS2020_15231a7c}.
Treating the model as a generic smooth function $f$ mapping an input $x$ to an output $y$, physical consistency dictates that this output must transform predictably under the symmetry group $G$ relevant to the system.
When equivariance is hard-coded into the architecture, these transformation laws are exact (see \autoref{fig:fig1}b): the energy of a molecule is invariant under global rotation, while velocities rotate covariantly with the reference frame. Formally, a function $f$ is equivariant with respect to a group $G$ if $f(gx) = \rho(g) f(x)$ for all $g \in G$, where $\rho$ is the representation of the group acting on the output space. For scalar targets (e.g., energy), $\rho(g)$ is the identity, recovering the definition of invariance: $f(gx) = f(x)$. Generally, if the output belongs to an irreducible representation (irrep) $\alpha$ of dimension $d_\alpha$, the group action is mediated by the matrix $\rho_\alpha(g)$.

When equivariance is learned from data rather than enforced, it is inherently approximate. To quantify the fidelity of a model $f$ to the transformation rules of a representation $\alpha$, we use the \emph{equivariance error} $A_\alpha$:
\begin{equation}\label{eq:equivariant-a}
    A_\alpha(f,x)=\sqrt{\left<\norm{f(hx) - \langle\rho_\alpha(g^{-1})f(ghx)\rangle_{g\in G}}_2^2\right>_{h\in G}},
\end{equation}
Here, angled brackets $\langle \cdot \rangle$ denote the Haar average over the group with respect to the transformations of the input for any continuous group, or a group average for discrete groups. 
Note that the term $\rho_\alpha(g^{-1})f(gx)$ represents the output transformed back to the reference frame of $x$; for a perfectly equivariant function, this quantity is constant (equal to $f(x)$). Therefore, $A_\alpha$ effectively measures the standard deviation of the re-projected outputs over the group orbit (\autoref{fig:fig1}c).
This metric has been already introduced in Ref.~\citenum{puny2022ICLR} over a carefully designed subset of group elements, later used in Ref.~\citenum{lang+24mlst} to give an estimate of the equivariance error over forces predicted by the Point-Edge Transformer (PET) MLIP model and, recently, in Ref.~\citenum{elhag2026arxiv} to quantify approximate equivariance in unconstrained models trained with a penalty term enforcing equivariance. In the SI we provide proofs that this is a proper metric of equivariance as it vanishes if and only if $f$ is exactly equivariant with respect to $\alpha$. Practically, as shown in the SI, this metric can be written as
\begin{equation}
    A_\alpha(f,x) = \sqrt{\langle \norm{f}_2^2 \rangle_G - \norm{\langle \rho_\alpha(g^{-1})f(gx)\rangle_{g\in G}}_2^2},
\end{equation}
which can be computed more efficiently, as it only requires a single group average and avoids the evaluation of the model $f$ over compositions of group elements. 

While the final output is optimized to be equivariant, the internal features of an unconstrained model are not bound to any specific symmetry constraints. Hidden layers may mix components of different equivariant characters while still effectively propagating information. To probe the symmetry content within the latent space, we introduce a spectral decomposition of the feature norm. 
Analogously to the power spectrum of a signal, we decompose the group-averaged norm of any internal feature $t$ (for example, the output of any hidden layer deep inside the model---see \autoref{fig:fig1}c) into contributions from different irreps:
\begin{equation}\label{eq:decomp-b}
\langle \norm{t}_2^2 \rangle_G = \sum_\alpha B_\alpha(t,x).
\end{equation}
Essentially, we look for a quantity that tell us the fraction of $t$ that transforms as the irrep $\alpha$ (more formally, the squared norm of the character-filtered components associated with the irrep $\alpha$, consistent with the Peter-Weyl decomposition\cite{folland_abstract_harmonic_analysis}).
The quantity of interest here is the \emph{character projection} $B_\alpha$:
\begin{equation}\label{eq:equivariant-b}
    B_\alpha(t,x)= d_\alpha^2\left<\norm{\left<\chi_\alpha(h^{-1}) t(hgx)\right>_{h\in G}}_2^2\right>_{g\in G},
\end{equation}
where $\chi_\alpha(g)=\operatorname{Tr} \rho_\alpha(g)$ are the group characters~\cite{VMK}. $B_\alpha(t, x)$, effectively defined in terms of group convolutions~\cite{Gerken2023,pmlr-v48-cohenc16,esteves2018polar}.
For convenience, we will use normalized character projections, $B_\alpha(t,x)/\langle \norm{t}_2^2 \rangle_G$, to simplify comparisons between different features and different layers of the architecture. Details about this quantity can be found in the SI, including an efficient evaluation scheme to minimize the required number of calls to the model function. We remark that other approaches~\cite{Gruver2023Lie} can give complementary information to the metrics introduced here~\footnote{We note that Lie-derivative-based diagnostics quantify equivariance error and, through the chain rule, admit a natural layer-wise decomposition of how this error is introduced across the architecture~\cite{Gruver2023Lie}. By contrast, the Haar-averaged quantities introduced here directly resolve the symmetry content of intermediate features and outputs in terms of irreducible representations of the full group. The two approaches are therefore complementary: one identifies where equivariance is broken, while the other reveals which symmetry channels are actually expressed. In the present context, this representation-resolved perspective is especially powerful, because it makes the emergence, suppression, and transfer of specific symmetry components directly visible throughout the model and along training.}.

Both $A_\alpha$ and $B_\alpha$ are independent of the choice of reference frame for the input $x$ (see the SI), which makes the two metrics intrinsic properties of the model and the input, rather than artifacts of a particular choice of orientation. 
Together, they provide complementary information on the symmetry of a model, with  $A_\alpha$ being appropriate to measure how accurately the outputs of the model obey the desired symmetry, and $B_\alpha$ being able to assess the representation characters that contaminate the outputs, and more broadly to identify the \emph{spectral content} of the model's internal features.

Although in this work we focus on $\mathrm{O(3)}$, the construction of $A_\alpha$ and $B_\alpha$ extends naturally to any compact group, including finite groups, by replacing the group average, irreducible representations, and characters with those of the symmetry group of interest. 
For compact groups such as $\mathrm{SU(3)}$, this generalization is therefore formal and direct. 
By contrast, for non-compact groups such as the Lorentz group $\mathrm{SO(1,3)}$, a normalized Haar average over the full group is not available, so practical counterparts of the present diagnostics would require restricted or task-dependent sampling measures over the group orbit.

\section{Unconstrained models for atomistic simulations}\label{sec:atomistic}

Surrogate models that target the outputs of quantum mechanical calculations of atomic-scale structures are an ideal starting point to demonstrate the conceptual and heuristic value of a symmetry analysis based on the introduced metrics.
These models' inputs are usually atomic positions and chemical species, formally described as \emph{decorated point clouds}. 
The point cloud edges are equivariant under any combination of global rotations, inversions, and translations, making the relevant group here the Euclidean group, E(3). 
Translational invariance is easily enforced by using interatomic distances rather than absolute positions, as is typical for graph-based models, so we can just focus on the orthogonal group O(3).

Physical properties targeted by these models naturally transform under (selected) irreps of O(3), and therefore unconstrained models must learn the appropriate transformation laws \textit{via} data augmentation.
We indicate the irreducible representations of the O(3) group by $(\lambda, \sigma)$, where $\lambda = 0, 1, 2,\ldots$ is the \emph{angular order}, and $\sigma = \pm 1$ the \emph{parity} under inversion, which discriminates between the proper subspace $(\sigma = +1)$ and its pseudo complement $(\sigma = -1)$. 
For example, scalars are proper invariants $(0, +1)$, vectors belong to $(1, +1)$, and pseudovectors to $(1, -1)$. In the SI, we provide a detailed discussion of the explicit form of the equivariance error and the character projections for this group. 
In practice, the required group averages are computed using a product integration grid (Lebedev~\cite{Lebedev1999AQF} on the sphere, plus trapezoidal over the remaining Euler angle~\cite{VMK,SHIMIZU2023108583}), which affords a computationally efficient and exact evaluation of the metrics introduced earlier (details and proofs in the SI).

We focus our analysis on the PET architecture, a transformer-based graph neural network (GNN) that takes both edge distances and vectors of an atomic-point cloud as inputs~\cite{pozd-ceri23nips}. 
It has demonstrated excellent performance as an MLIP in both standard benchmarks~\cite{pozd-ceri23nips,bigi2026+arxv} and materials science applications~\cite{mazitov_pet-mad_2025,Fadillah2025,turk2025}, and it serves as a representative example of a growing family of transformer-based architectures in this domain~\cite{kreiman2025transformersdiscovermolecularstructure,10.1063/5.0295035}. 
Data augmentation during training helps the model learn the O(3) invariance of the PES, leading to negligible equivariance errors compared to baseline accuracy, and to stable dynamics in the condensed phase~\cite{lang+24mlst}. 

We study the PET architecture in steps of increasing complexity to highlight different aspects of diagnostics and design.
We start in \autoref{sec:learn-pes} by diagnosing a universal MLIP trained on the MAD-1.5 dataset~\cite{MAD-1.5},  targeting the potential energy (scalar, $(0,+1)$ irrep), atomic forces (vectors, $(1,+1)$ irrep), and cell stress (symmetric, rank-2 Cartesian tensor, $(0,+1)$ and $(2,+1)$ irreps). 
Then, in \autoref{sec:black-box}, we track the spectral decomposition of angular information across various layers of the architecture, from random initialization through to the fully trained model. With the insights gained here, in \autoref{sec:purify}, we propose a simple post-hoc processing of the readout weights that further purifies the symmetry content of the outputs, leading to lower equivariance errors.
Finally, in sections \ref{sec:pseudo} and \ref{sec:high-lambda}, we investigate how the inductive biases built into PET affect learning dynamics for physical targets beyond the PES, and demonstrate how a symmetry analysis can inform simple but effective architectural modifications to improve learning dynamics and model accuracy.

\subsection{Learning the potential energy surface}\label{sec:learn-pes}

\begin{figure}[tb]
    \centering
    \includegraphics[width=\linewidth]{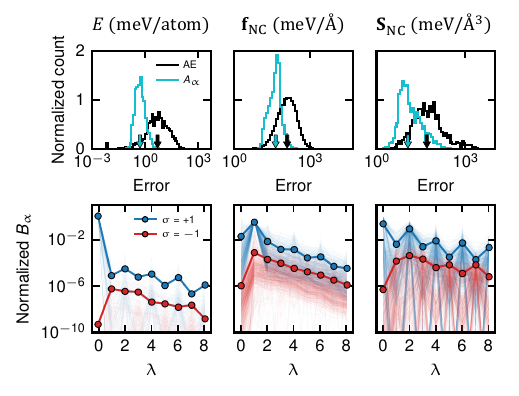}\vspace{-5mm}
    \caption{Equivariance diagnostics for a PET MLIP. Top: distributions of the absolute error (AE) and equivariance error, $A_\alpha$, for energy $E$, non-conservative forces $\mathbf{f}_{\text{NC}}$, and non-conservative stress $\mathbf{S}_{\text{NC}}$. The arrows on the x-axis indicate the distribution medians. Bottom: normalized character projections, $B_\alpha$, for the corresponding quantities as a function of the probed angular momentum channel $\lambda$. Solid lines and markers are averaged over 150 randomly sampled test structures, while faint lines show the individual structure projections.}
    \label{fig:fig2}
\end{figure}

We trained a PET MLIP on the 1.5 version~\cite{MAD-1.5} of the Massive Atomic Diversity (MAD) dataset~\cite{mazitov_massive_2025}. We train the model from scratch for 2000 epochs to properly examine its symmetry content at random initialization and track its evolution during training. We first evaluate the trained model on energy, forces, and stress of a subset of the MAD-1.5 test split. The top panel of \autoref{fig:fig2} shows a comparison of the resulting distribution of equivariance errors with that of the absolute model errors.
In all cases, the distribution of equivariance errors is shifted towards smaller values than that of the absolute errors, with the median of their ratios being 10\%, 31\%, and 26\% for energy, non-conservative forces, and non-conservative stress, respectively. We focus on forces and stresses predicted directly as outputs of the network to investigate the symmetry behavior of PET for higher order irreps, as when computing them as  derivatives of the energy the tensorial nature arises due to the differential operators and it is therefore directly tied to the invariant behavior of the energy.  

\begin{figure}[tb]
    \centering
    \includegraphics[width=\linewidth]{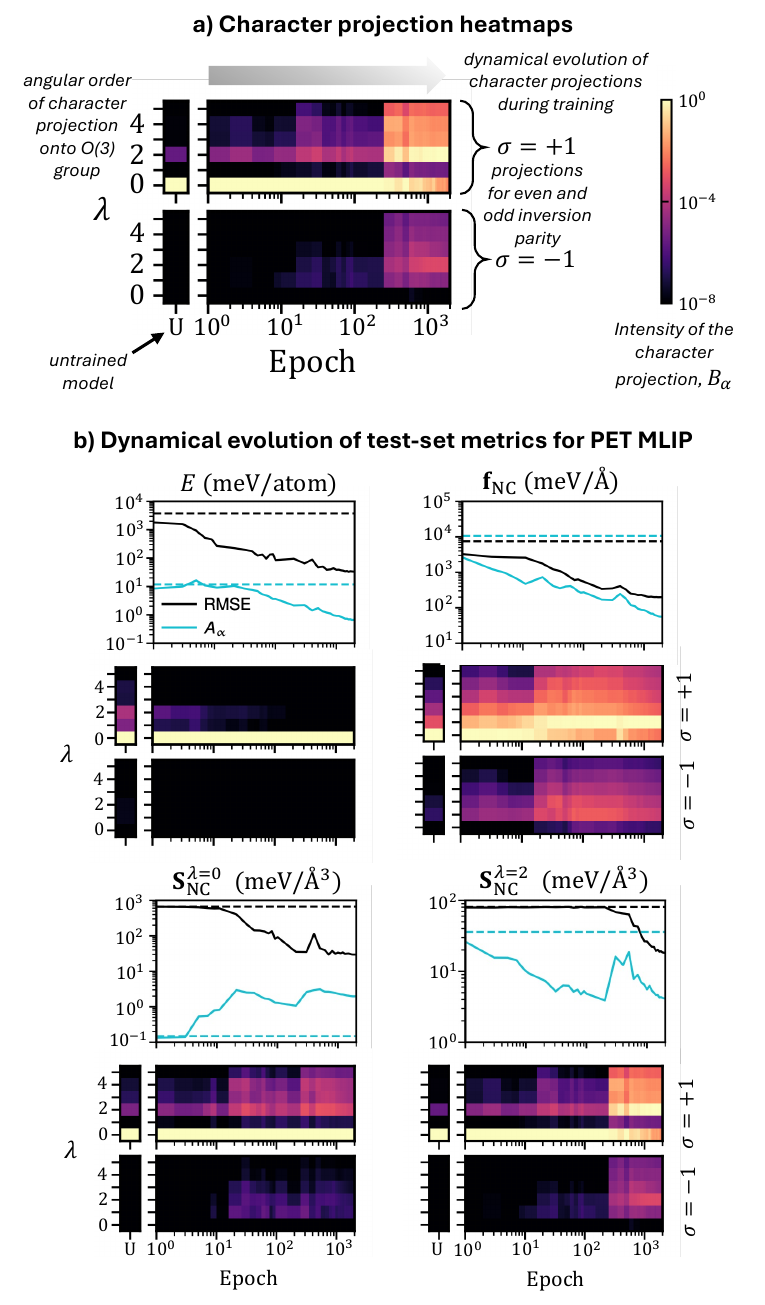}
    \caption{(a) Character projection heatmaps report on the magnitude of the character projection $B_\alpha$ for a quantity as a function of the character $\alpha$ of the relevant group and along successive epochs of a training run. In this case the characters are the $(\lambda, \sigma)$ irreps of the O(3) group. (b) Training curves and character heatmaps for energy, $E$, non-conservative forces, $\textbf{f}_{\text{NC}}$, and the two irreducible spherical components of the non-conservative stress, $\textbf{S}_{\text{NC}}$.}
    \label{fig:character-heatmaps}
\end{figure}

Given that the outputs are not \emph{exactly} equivariant, we can use the normalized character projections to investigate the nature of the errors (\autoref{fig:fig2}, bottom panel, showing $B_\alpha(y,x)$ for 150 randomly sampled test structures).
As expected, the scalar channel heavily dominates the energy predictions, with negligible contributions from higher angular-momentum or pseudo subspace channels~\footnote{The potential energy has a large ``compositional baseline'' built as a sum of species-dependent learned atomic energies. We have removed this trivially invariant term from the energy predictions before the computation of $B_\alpha$.}.
Forces manifest as vectors with active $\lambda = 1$ channels, while the stress tensor (a symmetric Cartesian tensor of rank 2) activates both both $\lambda = 0$ and $\lambda = 2$ proper channels~\cite{gris+18prl,domina2025JCP}. The second important point to note is that pseudo channels are always less active than proper ones. This is dramatically evident in the case of energy, while the difference is less marked for forces and stress. In particular, the pseudoscalar $(0, -1)$ channel carries very low signal intensity in all cases, and does not follow the exponentially decaying trend seen for the higher-angular order terms of the pseudo-subspace $(\lambda > 0, \sigma=-1)$.

\begin{figure*}[t]
    \centering
    \includegraphics[width=\linewidth]{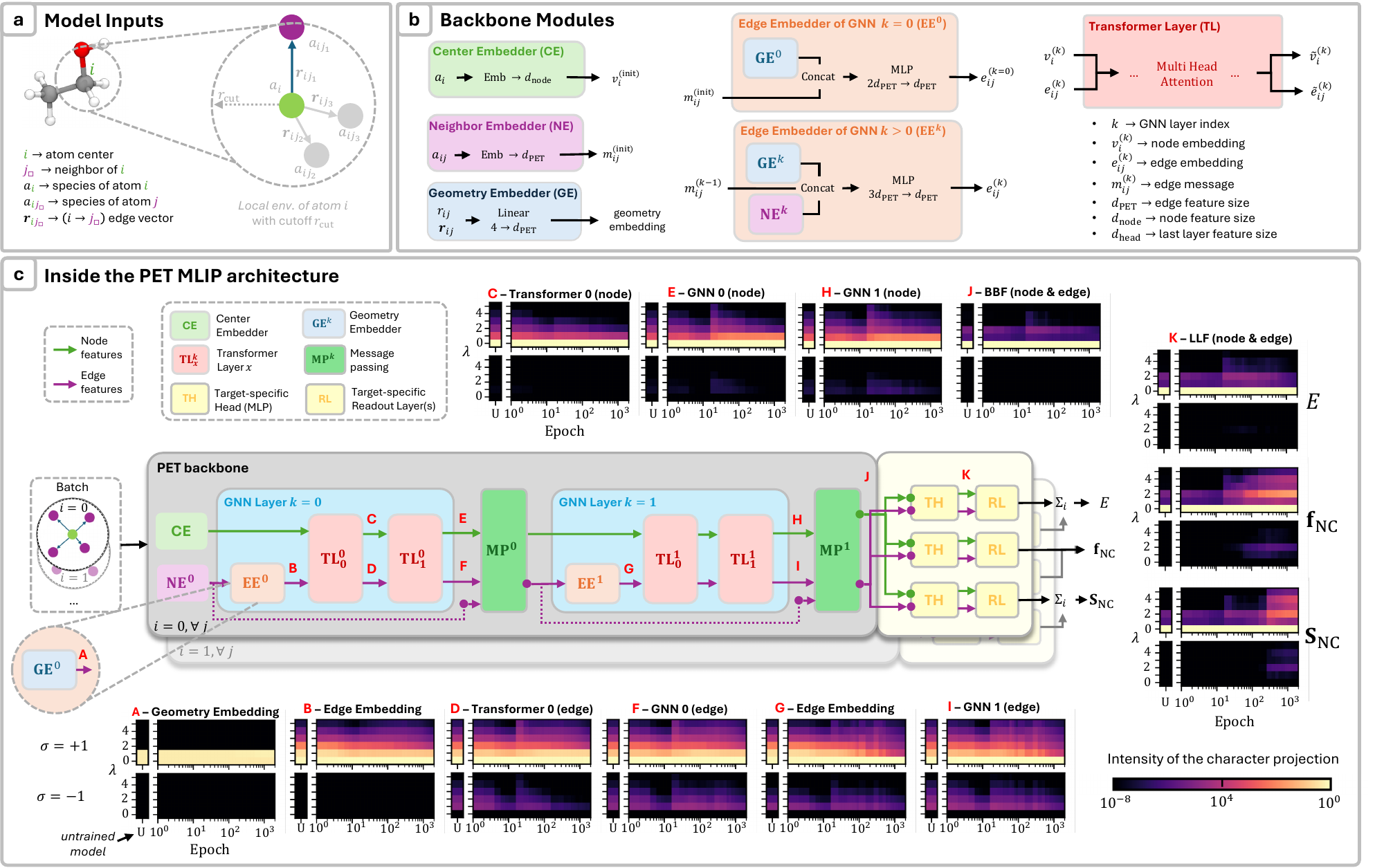}
    \caption{Overview of a PET MLIP architecture and dynamical evolution of the internal features. (a) Local atomic environments of atoms in a molecule/material are represented as decorated point clouds. The atomic species of the central atom ($a_i$) and its neighbors ($a_{ij}$), together with the edge vectors ($\mathbf{r}_{ij})$ and their magnitudes ($r_{ij}$), form the inputs to PET. (b) Backbone modules: embedding modules map the inputs to the latent space via Center, Neighbor, and Geometry Embedders. These feed into one Edge Embedder per GNN layer, which then enters the Transformer Layers. (c) Full architecture: the complete model pipeline leading to predictions for energy $E$, non-conservative forces  $\mathbf{f}_\text{NC}$, and non-conservative stress $\mathbf{S}_\text{NC}$. The following architecture hyperparameters were chosen: number of GNN layers $n_{\text{GNN}} = 2$, number of transformer layers $n_{\text{TL}} = 2$, cutoff $r_{\text{cut}} = 4.5$ \AA, edge feature size $d_{\text{PET}} = 128$, node feature size $d_{\text{node}} = 256$, LLF size $d_{\text{head}} = 128$. The surrounding heatmaps, labeled with red letters (A--L), correspond to specific points in the architecture. They display the intensity of the normalized character projections, $B_\alpha/\langle\|t\|_2^2\rangle_G$, as a function of the training epoch ($x$-axis, log-scale) and the probed $\lambda$ channel ($y$-axis). Within each heatmap group, the top and bottom panels represent the $\sigma$ channels, respectively (as detailed in heatmap A). The thin, isolated column on the far left of each heatmap represents the untrained model, denoted by ``U'' on the $x$-axis.}
    \label{fig:inside_black_box}
\end{figure*}

One can get more insights by monitoring how the characters evolve during a training run. To this end, we introduce character projection heatmaps~(\autoref{fig:character-heatmaps}a) to visualize the evolution of $B_\alpha$ for different angular momentum and parity channels, starting from the untrained, randomly initialized model all the way to the fully trained model. Dark colors indicate smaller components, and bright colors indicate larger components for each given channel. At any given vertical slice (i.e., epoch), the heatmaps visualize the spectral decomposition analogous to those seen in the bottom panels of \autoref{fig:fig2}.

For each of the outputs of the PET model along training, we monitor the test-set model error, equivariance error and character projection heatmaps (\autoref{fig:character-heatmaps}b).
One can see that randomly initialized model has a strong bias towards scalar $(0,+1)$ character and have near-zero spectral content for $\lambda>2$ and/or $\sigma=-1$. This also manifests in the predictions during the first $\sim 20$ epochs of training regardless of the specific target.  
Then we observe a sudden transition, that is most evident for the non-conservative force prediction: the scalar character drops, and the $(1,+1)$ vectorial character becomes dominant, accompanied by a sudden jump in high-$\lambda$, and $\sigma=-1$ components.
At the same point in the training run, the non-scalar errors in the energy output drop to almost zero.
During later epochs, as both the target RMSE and equivariance errors decrease, the character decomposition becomes more sharply peaked around the desired irreps, without other sudden transitions apart from the appearance of the $\lambda=2$ stress component at epoch $\sim 200$. The scalar stress component is the only target whose equivariance error increases during training from its initial value: the reason is that at the beginning of the training the scalar stress predictions are close to zero, which is trivially invariant. A similar increase is seen, for a similar reason, for the $\lambda=2$ component of the stress. The late onset of learning for the non-conservative stress implies that the model is not fully converged at the end of the run, with consequences that will be apparent in Section~\ref{sec:purify}.

\subsection{Looking into the black box}\label{sec:black-box}

The character decomposition analysis of PET demonstrates that there is a lot of structure in the symmetry-breaking errors, and that the model performs especially well on invariant targets. To understand \emph{why}, and to how to make PET better at learning targets that are not invariant, we have to look deeper at \emph{how} it learns symmetry.
To do so, we explore the information content of PET features on the two aforementioned axes: across layers of the architecture and throughout the training process.

\autoref{fig:inside_black_box} introduces a high-level overview of the PET MLIP architecture. For the outputs of several important hidden layers of the model trained in the previous section, we perform character decomposition analysis, with the resulting heatmaps annotating the architecture. The atom-centered environments of a molecule or material form the input to the architecture as decorated point clouds (\autoref{fig:inside_black_box}a). Internal representations of these environments are generated by the backbone PET architecture. This is comprised of a series of message-passing GNN layers, each of which contains several modules (\autoref{fig:inside_black_box}b). The resulting backbone features (BBF) contain rich geometric information common to all model outputs. These are transformed into last-layer features (LLF) by target-specific nonlinear heads, which are mapped to model outputs by linear readout layers. At the coarsest level, the internal representations constructed by the backbone architecture evolve with rich dynamics during training, developing high-$\lambda$ as well as pseudotensorial ($\sigma=-1$) characters, which are especially strong for the edge features.

In analyzing the finer details of the backbone architecture, we can see how the character trends we observed for the model outputs develop. For a given local atomic environment in the batch, the chemical nature of the central atom and its neighbors are embedded as learnable scalar tokens (center embeddings (CE) and neighbor embeddings (NE)), with geometry embeddings (GE) representing the local 3D structure---specifically, the distances $r_{ij}$ and interatomic vectors $\mathbf{r}_{ij}$ to each neighbor (\autoref{fig:inside_black_box}b).
These GE have pure scalar and vectorial nature, which does not change during training (heatmaps A). A multi-layer perceptron converts the GE into fixed-size edge embeddings (EE). The nonlinearity generates higher-$\lambda$ characters, but still lacks pseudo-tensorial character (heatmaps B). 

The transformer blocks, at the heart of the PET architecture, are capable of generating rich geometric features thanks to the attention mechanism, which mixes information from multiple edges and enables activation of pseudo-tensorial channels (heatmaps C--F). This is especially true for edge features, though the signal still remains weak.
After two transformer layers, the first GNN layer is complete, and at this point the model has been able to combine geometric information within a single neighborhood. Even at initialization, with random weights, the outputs of the GNN block are dominated by the scalar character, but contain significant higher-$\lambda$ components (heatmaps E, F).
The message passing step combines geometric information with edge messages from the previous step. As such, the pseudotensorial activation gained in the first GNN layer propagates to the next and is therefore visible in the new EE (heatmaps G), while no other notable changes in the outputs of the second GNN layer (heatmaps H, I) compared to the first exist.
The BBF are obtained after aggregation of the outputs of all GNN layers. Interestingly, this aggregation suppresses $\lambda > 2$ and all $\sigma = - 1$ channels, with a strong surviving scalar expression.

The BFF are then split between several target-specific nonlinear heads and mapped to the LLF (heatmaps K). These are the last nonlinear operations in the network. In the randomly initialized model the LLF have only low-order character, but at later stages they develop a richer geometric structure, that persists until the end of the training. This can be seen particularly for the forces.
Finally, the LLF are mapped to predictions via the linear readout layers. These can only modulate the magnitude of the existing components, and act as a filter that determines the final characters and equivariance errors of the predictions, as seen previously in \autoref{fig:character-heatmaps}.

This detailed analysis provides a compelling picture of how PET learns to be (approximately) equivariant. It also gives indications that could be useful for the design of equivariant models, because the character decomposition learned by the unconstrained architecture suggests what types of symmetry channels are needed to learn efficiently energy, forces and stresses. 
It shows that, with the GE we use, PET is strongly biased towards low-$\lambda$ components, and that even if the network has the expressive power to generate higher-order character features, it uses this power sparingly, converging to intermediate representations dominated by low-$\lambda$, $\sigma=+1$ character. 
This is consistent with the observation that equivariant networks can often yield reasonably accurate predictions with internal representations capped at $\lambda_\text{max}=2$~\cite{bazt+22ncomm}. 
The presence of high-order components in the edge representations suggests however that a higher resolution is important to process geometric message-passing information, so it might be useful to experiment with equivariant models that invest some computational budget into including higher-$\lambda$ terms, at least for some of the GNN layers.

\subsection{Symmetry purification of the readout}\label{sec:purify}

In the previous section we observed that the final readout layers must filter out undesired symmetry content from the LLF. 
This is achieved through symmetry augmentation with random rotations during training, resulting in a model with equivariance error that is a small fraction of the model error. 
Still, the LLF tokens are contaminated by a significant fraction of irreps not needed to describe the target subspace.
We propose here a simple protocol to purify the linear readout, which can be formulated as a regularized regression problem.
If $\phi(x)$ are the LLF for the input $x$, the linear readout can be expressed as $y(x)=\theta^T\phi(x)$. 
By computing the LLF over a group orbit (or a grid, in the case of continuous groups such as $O(3)$) we can define two loss terms,
\begin{equation}
\begin{split}
L_\mu = &\left\langle \left\|\rho(g^{-1})\theta^T\phi(g x)- y\right\|^2_2 \right\rangle_{g \in G},\\ 
L_\sigma=&
 \left\langle\|\theta^T\phi(g x)\|^2_2 \right\rangle_{g \in G} -\| \left\langle 
\rho(g^{-1})\theta^T\phi(g x) \right\rangle_{g \in G}
 \|^2_2.
\end{split}
\end{equation}
The first describes the model mean-square error averaged over the group, while the second corresponds to the squared equivariance error $A$. 
The combined loss $L=L_\mu + \gamma L_\sigma$, where $\gamma$ controls the relative weighting of the equivariance penalty. can be cast into an explicit least-squares problem, and solved for the readout weights $\theta$. 
All the terms needed to determine $\theta$ can be computed with a single sweep over the training set. 
The derivation of a closed form for the optimal weights that minimize this loss is reported in the SI.

\begin{figure}
    \centering
    \includegraphics[width=\linewidth]{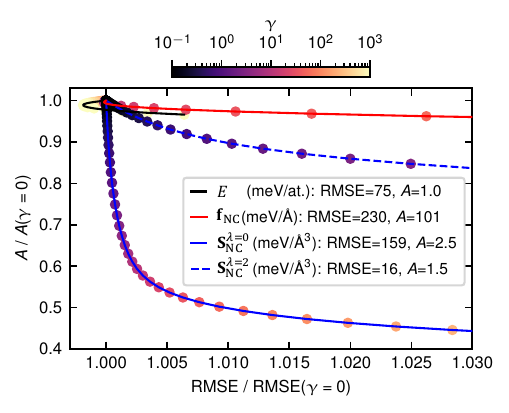}
    \caption{Model error and equivariance error for the energy $E$, non-conservative forces $\textbf{f}_\text{NC}$ and $\lambda=0,2$ components of the non-conservative stress $\textbf{S}_\text{NC}$ for a universal PET model (trained and tested on MAD-1.5) whose energy and non-conservative force readout layers have been retrained with a loss $L=L_\mu + \gamma L_\sigma$ combining model error and an equivariance error penalty. The marker colors correspond to the weighting of the equivariance penalty, $\gamma$.}
    \label{fig:retrain_ll}
\end{figure}

For the PET model we find that this procedure yields a modest improvement of the equivariance error for most of the outputs, confirming that on-the-fly augmentation performed during conventional model training succeeds in minimizing $A$. 
The exception is the conservative stress, for which we noted that the proper symmetry channels became active only towards the end of the training. In this case, by  tuning the weight of $L_\sigma$, the equivariance error can be reduced by a factor of 2 sacrificing less than 1\%{} in the RMSE relative to the target. 
The inclusion of an explicit equivariance penalty during primary training, as in Ref.~\citenum{elhag2026arxiv}  would also influence the optimization of the backbone weights, but requires multiple model evaluations per structure, increasing the computational cost of training.
Instead, the equivariant readout optimization we propose here has a cost equivalent to running symmetrized inference only once on each structure in the dataset, and can be applied routinely as a way to validate, and possibly improve, the symmetry properties of unconstrained models.

\subsection{Stress-testing the geometric expressivity of the model}
\label{sec:pseudo}

\begin{figure}
    \centering
    \includegraphics[width=\linewidth]{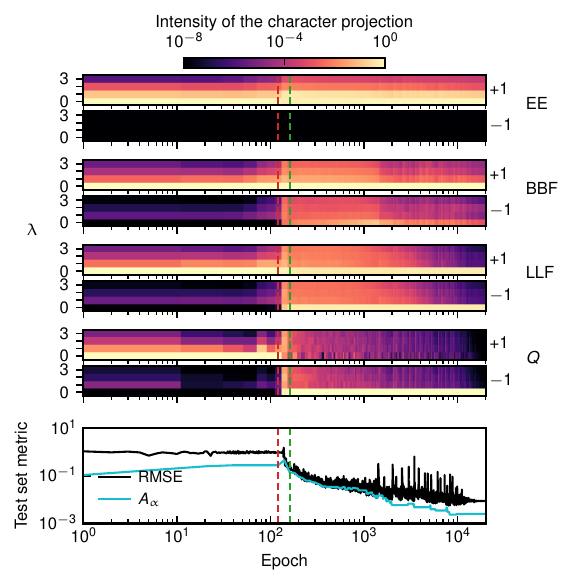}\vspace{-4mm}
    \caption{Equivariance diagnostic for the geometric pseudoscalar training. Top: Character projection heatmaps for a PET model across architectural layers over the course of training. Bottom: Test set RMSE and average equivariance error. The region between the red and the green dashed lines marks the onset of rapid learning. This phase is driven by a strong activation of the pseudoscalar channel across all capable layers, accompanied by a broader, weaker activation of higher-order tensorial ($\sigma=+1$) and pseudo-tensorial ($\sigma=-1$) channels (indicated on the right).}
    \label{fig:fig4}
\end{figure}

The fact that the internal representations are dominated by low-order characters raises the question of whether PET can learn targets with a symmetry that is not naturally expressed by the initial random weights, which we are going to investigate next. 
The clearest signal from the symmetry analysis of the model features is that $\sigma=-1$ irreps, and pseudoscalars in particular, have negligible character throughout the PET architecture.
To determine whether this impacts the representation power of the model, we define a purely geometric pseudoscalar target $Q$ as the the triple products of all triplets of bonds in a molecular system:
\begin{equation}\label{eq:Q-definition}
    Q = \sum_{i<j<k<l} (\mathbf{r}_i-\mathbf{r}_l) \cdot \left[ (\mathbf{r}_j-\mathbf{r}_l) \times (\mathbf{r}_k-\mathbf{r}_j) \right],
\end{equation}
and generated a toy dataset of 1000 rattled conformers of bromochlorofluoromethane (CHBrClF) differing only in initial orientation and small random coordinate distortions. We trained PET from scratch using an 80:10:10 split.

The top panels of \autoref{fig:fig4} show the character projection heatmaps for the internal features (EE, BBF, LLF) and predictions ($Q$) over the training epochs, while the bottom panel reports the RMSE and equivariance error ($A_\alpha)$. 
The network exhibits a striking two-phase learning behavior along the training, similar to the so-called grokking phenomenon often observed in large-language models~\cite{grokking}---albeit in this case on both training and validation sets.
From this point of view, this observation suggests that the ability of the network to generate high-order symmetry terms starting from simple vectorial descriptors might be interpreted as a phase transition in the parameter space of the transformer~\cite{liu2022understandinggrokkingeffectivetheory}.
During the initial phase (up to $\sim 120$ epochs, red dashed line), only proper ($\sigma=+1$) channels are active, leaving pseudo features virtually dormant. Because the proper (and especially the purely invariant) channels lack the necessary symmetry to express $Q$, both the test RMSE and $A_\alpha$ stagnate. 
A dramatic shift occurs between the red and green dashed lines, where pseudotensorial channels suddenly activate across the BBF, LLF and output ($Q$). This internal transition triggers a steep and continuous drop in both RMSE and $A_\alpha$.
As expected, the EE features show no pseudotensorial activation because at this stage---just before the transformer---the model processes edges individually and cannot yet combine them.  
Constructing a pseudotensor requires mixing at least two vectors (a ``second-order'' effect), while forming a pseudoscalar requires combining at least three (a ``third-order'' effect, see the SI for details). 
The network must synthesize these pseudo representations from strictly proper inputs within the attention layer, leading to a delayed onset of learning. Because pseudo channels are initially weak, their gradient signals are suppressed and their construction requires coordinated higher-order interactions across multiple features, making optimization intrinsically slow until a threshold is reached where these components can self-amplify.
Given that $Q$ is a purely geometric target, somewhat trivially defined from molecular geometry, it is remarkable that even in this case the learning task is so difficult.

This diagnostic shines light on how architectural design restricts the model's ability to build features with a specific group symmetry and how this may impact learning dynamics. 
Under-representation of the pseudo subspace carries implications beyond artificial tasks, potentially impacting real-world atomistic applications, such as learning of NMR chemical shielding tensors~\cite{kellner2025JPCL} (which require pseudo-vectorial irreps), circular dichroism~\cite{Zhao2021} or targeting Hamiltonian and density matrices that contains many components with $\sigma=-1$ character~\cite{niga+22jcp,Li2022,suma+25jctc,Febrer2025}.

\subsection{Increasing the geometric expressivity of PET}
\label{sec:high-lambda}

\begin{figure}
    \centering
    \includegraphics[width=\linewidth]{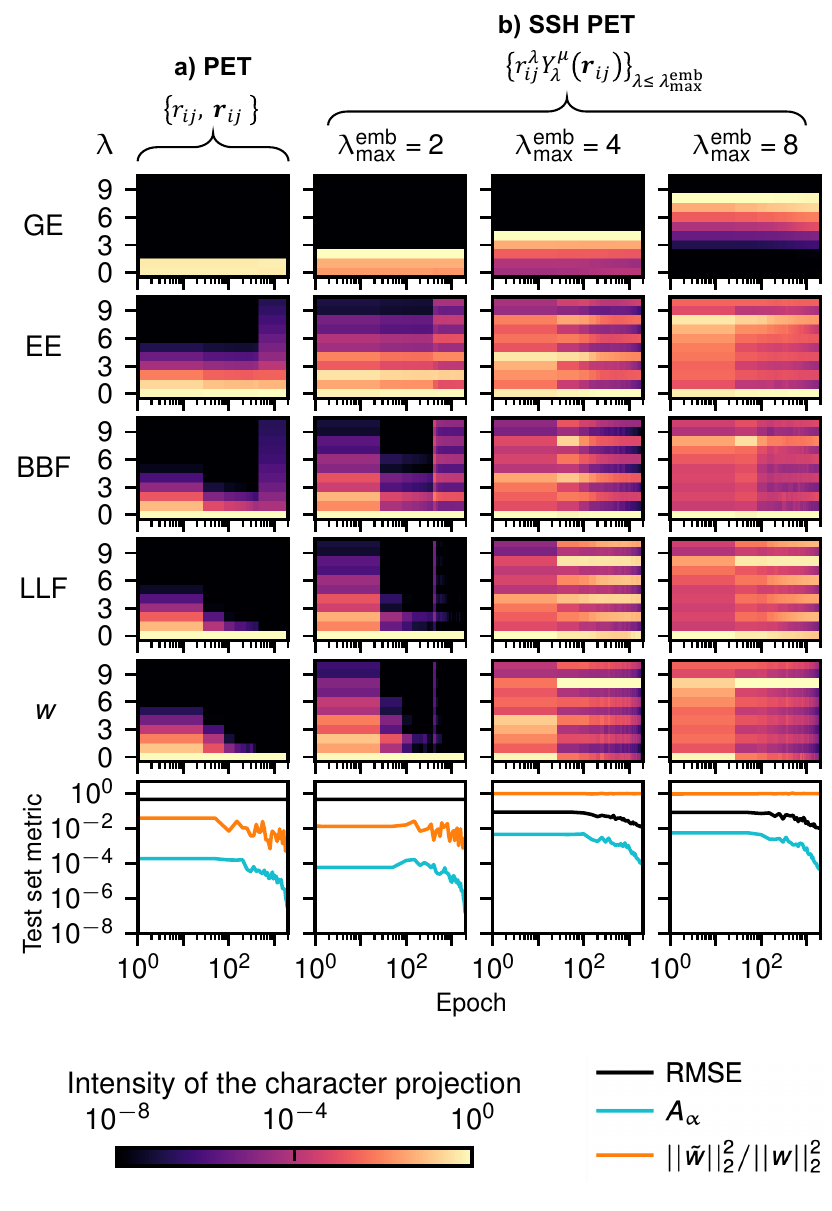}\vspace{-4mm}
    \caption{Impact of higher-angular order inductive biases on learning high-angular order targets. Standard PET (first column) and modified variants---which replace standard embeddings with solid spherical harmonics up to $\lambda_{\text{max}}^{\text{emb}} \in \{2, 4, 8\}$---are trained on the $\lambda_{\text{target}}=8$ channel of the electron density projections ($w$).
    Top: proper-tensorial ($\sigma=+1$) heatmaps tracking the angular order ($\lambda \in \{0, ..., 10\}$) evolution of features across key network layers during training. Colors denote the normalized intensity of the character projection. Bottom: validation metrics per epoch, including prediction RMSE (black), equivariance error (cyan), and the predicted-to-target norm ratio (orange).}
    \label{fig:high-L}
\end{figure}

Next, we challenge the PET model to learn a high-$\lambda$ target. For this we start with the real-space electron densities, $\rho(\textbf{r})$ (defined on a set of grid points, $\textbf{r}$) for the spin-singlet, charge neutral, organic (covering 11 elements) subset of the QCML molecular dataset~\cite{QCML-dataset} computed with FHI-aims\cite{FHI-aims-2025}. The densities are then projected onto an atom-centered auxiliary basis set (a product of spherical harmonics and radial functions), $\varphi_{in\lambda\mu}$, to yield the \emph{electron density projections}, $\mathbf{w}$:
\begin{equation}
    \text{w}_{b} = \langle \varphi_{b} (\textbf{r}) | \rho (\textbf{r}) \rangle \ ,
\end{equation}
where $b \equiv \{in\lambda\mu\}$ is a composite index labeling the atom $i$, radial channel $n$, angular order $\lambda$, and component $\mu$. The auxiliary basis set is chosen to be expressive enough to expand the electron density with low basis-set expansion error as $\rho(\textbf{r}) \approx \sum_{b} c_{b} \varphi_{b} (\textbf{r})$, with $c_{b'} = \sum_b S_{bb'}^{-1} \text{w}_b$ being the expansion coefficients and $S$ the auxiliary basis overlap matrix. The electron density projections form a challenging target derived from a quantum mechanical observable that is central in atomistic modeling and, crucially, decompose onto a series of proper-tensorial irreps up to some maximum angular order, $\lambda_{\text{max}}^{\text{target}} = 8$ in our case. By isolating and targeting only the highest-$\lambda$ channel, we can observe clearly how the absence of high-angular order features at initialization affects learning in a realistic scenario.

\autoref{fig:high-L}a tracks the architecture's attempt to learn the $(8, +1)$ components of the density projections.
As shown by the entirely flat RMSE curve over 2000 epochs, standard PET fails to learn this target. Because the network only embeds edge distances and vectors, the initial GE strictly activate at $\lambda=\{0,1\}$. While the nonlinear multi-layer perceptron generates some higher-$\lambda$ activation in the EE, it is insufficient. Lacking meaningful $\lambda=8$ information in its features, the model  collapses to a trivial solution: it predicts zeros to minimize the loss, as evidenced by the declining equivariance error and a vanishing norm ratio between predictions and targets. 

\begin{figure*}[th!]
    \centering\includegraphics[width=0.9\linewidth]{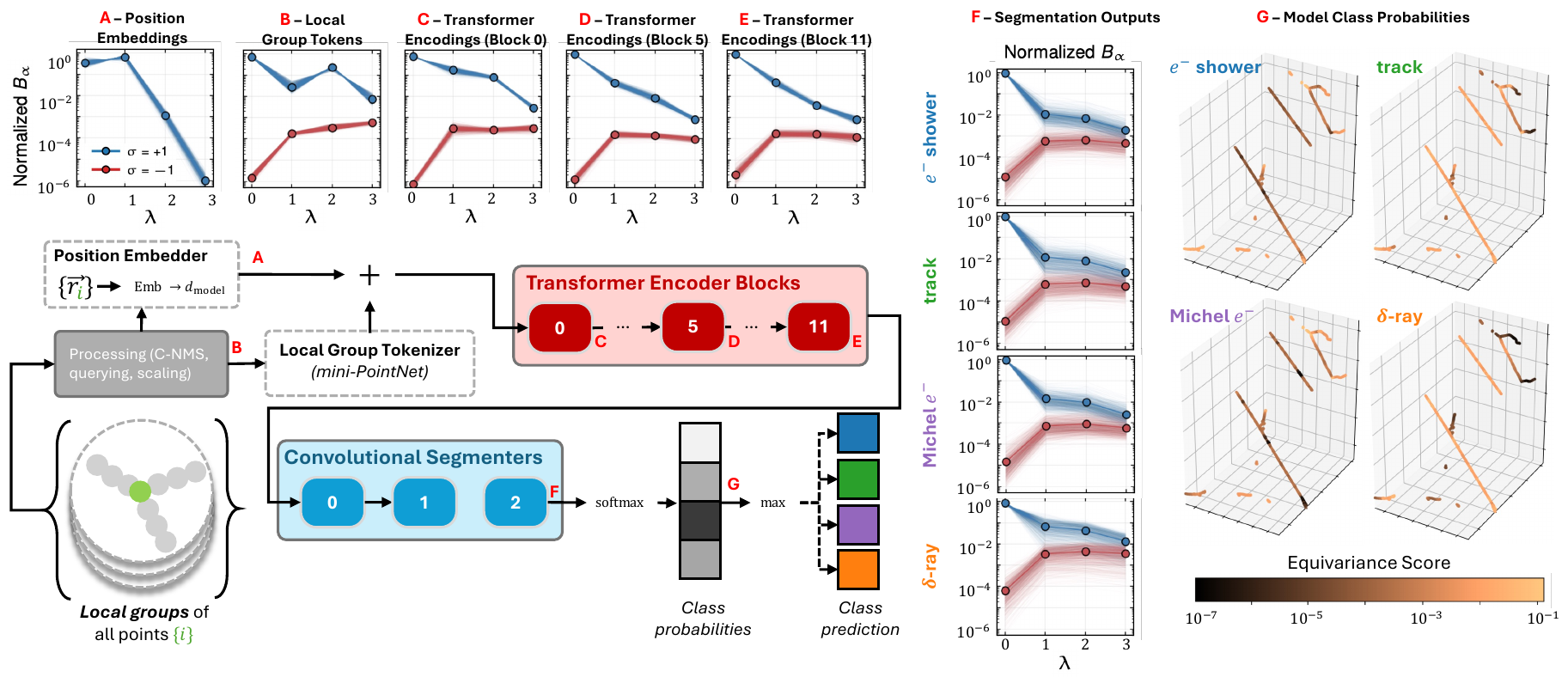}
    \caption{Schematic representation of the PoLAr-MAE architecture, with the character decomposition of the internal tokens and the segmentation outputs plotted for 1000 test events. For representation purposes we show the decomposition after aggregation over the points of each event. Equivariance errors for the model class probabilities are shown for a representative event (the same as in \autoref{fig:polarmae-overview}), showing how values of $A$ correlate either with branching points or trajectory segments for which the classification is unstable with respect to rigid rotations of the reference frame.}
    \label{fig:polarmae}
\end{figure*}

\begin{figure}
\centering\includegraphics[width=0.95\linewidth]{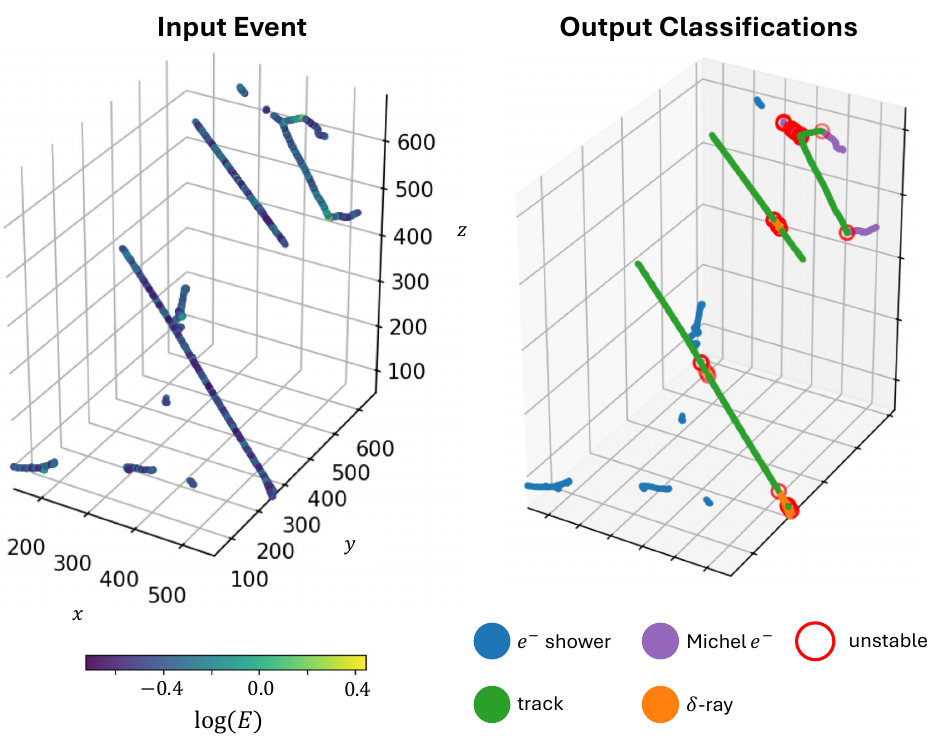}
\caption{
A representative input and output of the PoLAr-MAE model. For a given input event (left panel), e.g., clusters of particle tracks (point clouds) in a liquid argon detector decorated with particle energies, the model makes point-wise classifications (right panel) of the discrete event class. Points surrounded by a red circle change classification when the structure is rigidly rotated. Axis labels are given in terms of voxel indexes. 
}
    \label{fig:polarmae-overview}
\end{figure}

We propose to address this deficiency in the architecture by a simple and inexpensive modification to the edge geometry embeddings. We replace standard edge distances and vectors with a solid spherical harmonics (SSH) expansion of the edge vectors up to a maximum angular order, $\lambda_{\text{max}}^{\text{emb}}$. 
SSHs naturally represent both angular structure and radial information as each carries a well-defined angular momentum channel by construction, and can be evaluated in an auto-differentiable manner with negligible overhead~\cite{bigi+23jcp}. 
In fact, due to the functional form of the SSHs, $r_{ij}^{\lambda} Y_{\lambda}^{\mu} (\mathbf{r}_{ij})$, the GE of standard PET is almost equivalent in information content to that of an (embedded) SSH expansion capped at $\lambda_{\text{max}}^{\text{emb}} = 1$. By increasing this cap, we provide the model with a vastly superior starting point to describe high-$\lambda$ quantities.

To investigate the effect of this modified inductive bias, we trained from scratch the same PET architecture as in \autoref{fig:high-L}a, modifying only the GE layer to use SSH embeddings of increasing maximum angular order $\lambda_{\text{max}}^{\text{emb}} \in \{2, 4, 8\}$. As shown in \autoref{fig:high-L}b, explicitly increasing the input's angular information drastically alters the learning dynamics. While $\lambda_\mathrm{max}^\mathrm{emb}=2$ behaves similarly to standard PET and fails to learn, models initialized with $\lambda_\mathrm{max}^\mathrm{emb}=4$ and 8 succeed. For these models, $\lambda=8$ character is visible in $B_\alpha$ right from the start, the RMSE steadily declines, and the norm ratio stabilizes around 1, proving the model is genuinely learning the target rather than collapsing to zero. By the end of the training, the internal features possess rich, high-order angular information, successfully adapting the implicit inductive biases of the architecture and the geometric nature of the physical target.

\section{Beyond atomistic simulations}\label{sec:particle-phys}

Even though we focus our analysis on the application of unconstrained ML models to atomistic simulations, there are many domains of physics that rely on models that do not obey exactly all the relevant symmetries. 
As a brief but instructive example, we consider the classification of particle trajectories in liquid argon time projection chambers (LArTPCs) neutrino detection experiments. We investigate the symmetry properties of the point-based liquid argon masked autoencoder (PoLAr-MAE) method~\cite{Young_2026}, which relies on an unconstrained architecture similar to the general-purpose PointNet~\cite{qi+17ieee} and is built upon the publicly available PILArNet dataset~\cite{adams2020pilarnetpublicdatasetparticle}. 
Particle trajectories are represented as a point cloud, with each point corresponding to a detector interaction, decorated by the energy of the detected particle. Based on this input, the model classifies points along a particle trajectory into four classes: \emph{electron shower}, \emph{track}, \emph{Michel electron}, \emph{$\delta$-ray} (cf.~\autoref{fig:polarmae}).
PoLAr-MAE shows overall excellent classification accuracy, except for a few fine-grained structures. 
We observe that, even though the classification is usually invariant to rigid rotations of the reference frame (the model is trained with random rotational augmentation, similar to PET) there are a few track segments for which the class becomes dependent on orientation, especially for the sub-token structures where the classification is unreliable in the first place.

In \autoref{fig:polarmae} we show a summary of the architecture of the model and in-depth diagnostics of the internal representation. 
In the figure we also show the point-wise equivariance error of the class probabilities for a representative event (the same depicted in \autoref{fig:polarmae-overview}). A point-wise character decomposition of the internal descriptors for this same event  is reported in the SI.
One sees clearly that the uncertainty in the classification is reflected in the equivariance errors for the four channels predicted by PoLAr-MAE. The presence of a substantial non-scalar behavior is already apparent from the character decomposition of the segmentation outputs (heatmap F) with a dominant scalar nature and exponentially decaying $\lambda$ characters (which are however pronounced for the $\delta$-ray output. The pseudo-characters are all much smaller as it was the case for PET and also in this case the pseudoscalar contribution is minuscule. 

Looking at the internal layers, we see that the position embeddings A have predominantly vectorial character (i.e. the position embedder does not generate high-order characters), whereas the local group tokens, that are processed by a mini-PointNet unit, display a richer structure, with a substantial $\lambda=2$ component (heatmap B).
The transformer encoder blocks (heatmaps C-E) don't display remarkable trends: the dominant character is scalar, and successive blocks progressively dampen the higher-$\lambda$ characters. However, the convolutional segmenters increase the contamination from non-scalar characters. 
Overall, PoLAr-MAE shows trends that are broadly similar to those observed for node features in PET, missing however the richer, high-$\lambda$ characters of PET's edge features, suggesting that introducing explicit edge representations might increase the expressive power of the model. 
Large equivariance errors are associated with instabilities of the classification. While introducing inference-time rotational augmentation could reduce the errors, a more elegant and computationally efficient solution would be to apply the readout purification protocol introduced in \autoref{sec:purify} to the final convolutional segmenters. By explicitly penalizing the equivariance error of the classification head over the $O(3)$ group, one could enforce rotational invariance for these challenging sub-token structures without incurring inference-time overhead.

\section{Conclusions}

Many fields of study, from image recognition to natural language processing, have come to accept the \emph{bitter lesson} that incorporating domain-field knowledge into the design of machine-learning models does not pay off, in the long run, against architectures designed to be simple, fast, and able to scale to large datasets. 
For a relatively long time, physics appeared to be the exception, with considerable effort being devoted, with excellent results, to the design of symmetry-adapted models that are bound to fulfill invariant or covariant behavior with respect to symmetry operations by design.
The mounting evidence that this is not, at the very least, \emph{necessary} is at odds with the foundational role played by symmetry, prompting the questions \emph{how}, and \emph{how accurately}, models without these built-in constraints come to learn symmetry. 
We develop an analysis framework that allows one to follow the character of the outputs as well as the internal representation of the model across the architecture and during a training run, and that is useful across domains, as we demonstrate by applying it to rotational symmetry in both atomic-scale simulations and the classification of particle trajectories from LArTPCs experiments.
Our analysis across domains suggest that ML models can learn rotational symmetry to a high accuracy, with symmetry errors being much smaller than model error. In fact, the two types of error are usually highly correlated.

We propose an inexpensive purification step for the linear readout of the model. We find that the PET-based atomistic model we consider is already trained to an excellent level using random augmentation, and that the improvements from this simple post-processing step are modest except for the scalar component of the stress for which the equivariance error can be halved.
The group characters of the internal representations of the model display highly non-trivial dynamics during training. Initially, low-angular-momentum $\lambda$ character dominates across the network, which we trace in part to the initial encoding of the geometry in terms of scalars and vectors. 
Higher-$\lambda$ terms appear soon for the internal representation, especially for the tokens associated with the inter-atomic edges, and only towards the end of the run the target properties achieve an accurate approximation of the desired symmetries.
The stress, that being a symmetric 3-tensor contains a $\lambda=2$ component, is learned only in this later stage.
This observation prompted us to investigate the performance of the model when learning pseudo-scalar and high-$\lambda$ target properties, associated with characters that are poorly represented in the network. We find that indeed the standard PET architecture struggles to learn these symmetries---a deficiency that we address by injecting higher-$\lambda$ descriptors at the level of the initial structural encoding, demonstrating the heuristic power of our symmetry analysis. 

The analysis of the particle trace classification model is broadly similar. The final features show a high degree of rotational invariance, and equivariance error correlates strongly with instabilities in the classification---indicating the importance of training the model to a high level of symmetry. 
Internal features show a predominantly scalar character throughout the model, which appears beneficial at least when performing a task that is based on invariant outputs.

A character analysis such as the one we introduce helps to select between different architectures and architectural hyperparameters, and to diagnose difficulties in learning symmetry behavior.
Some of the insights we extract might also be relevant for models that are equivariant by design. The fast decay of learned representation characters in the PET architecture validates the common choice of limiting the angular momentum order to $\lambda=1$ or $\lambda=2$ in the internal representation of equivariant MLIPs. 
When using the same architectures for high-$\lambda$ targets, however, it might not be enough to increase the order at the level of the heads, as we observe that lack of some components early in the network negatively affects the accuracy of outputs of the corresponding symmetry, despite the fact that the nonlinear heads we use can in principle generate terms of arbitrary order. 

The tension between general-purpose, flexible, and well-scaling architectures, and those that incorporate exact physical priors is unlikely to be resolved soon. 
In the meantime, the analysis framework introduced here provides a rigorous approach to quantify the symmetry behavior of ML models, yielding actionable insights that improve the accuracy and training dynamics of unconstrained architectures across the physical sciences. 
As demonstrated by our architectural modifications for learning the ground-state electron density, understanding a model's  failure modes allows researchers to achieve stability and accuracy by injecting the minimum required inductive biases, without sacrificing the expressivity and scalability of unconstrained networks.

\section*{Code availability}
All software components used in this study are open-source and freely available. 
The Python package used to implement and train our ML models is \texttt{metatrain}, on GitHub at \url{https://github.com/metatensor/metatrain}.
The complete set of data and workflows required to reproduce all figures in this manuscript is provided in a Materials Cloud~\cite{tarl+20sd} repository [DOI to be included when available]. 
\begin{acknowledgments}
The authors would like to ackowledge support by the following grants. 
MD and MC: the Swiss National Science Foundation (grant ID 200020\_214879). 
JWA and MC: the ERC Horizon 2020 Grant No. 101001890-FIAMMA.
FB: the Swiss Platform for Advanced Scientific Computing (PASC) and the Swiss AI Initiative (2025 Fellowship Program).
PP: a Sinergia grant of the SNSF (grant ID CRSII5\_202296).
Computation for this work relied on resources from the Swiss National Supercomputing Centre (CSCS) under the lp133 project, and the EPFL HPC platform (SCITAS). \end{acknowledgments}

\bibliographystyle{apsrev4-2}
\bibliography{biblio}

\onecolumngrid
\clearpage


\section*{Supplementary material}

# Supplementary Information for "*How unconstrained machine-learning models learn physical symmetries*"

M. Domina,[1, *] J. W. Abbott,[1, *] P. Pegolo,[1, *] F. Bigi,[1] and M. Ceriotti[1]

[1]*Laboratory of Computational Science and Modeling, Institut des Matériaux,*
*École Polytechnique Fédérale de Lausanne, 1015 Lausanne, Switzerland*
(Dated: March 25, 2026)

## I. EQUIVARIANCE ERROR

In this Supplementary Information we work with compact groups; the finite-group case follows by replacing Haar integrals with normalized sums over the group elements. We assume that the representations are taken to be unitary (orthogonal in the real case) which is always possible for compact groups. Throughout, $\mu$ denotes the normalized Haar measure on $G$, which is bi-invariant, so that $\mathrm{d}\mu(gh) = \mathrm{d}\mu(hg) = \mathrm{d}\mu(g)$ for all $g, h \in G$. For compact groups the Haar measure is also invariant under inversion, i.e. $\mathrm{d}\mu(g^{-1}) = \mathrm{d}\mu(g)$ for all $g \in G$. As done in the main text, $d_\alpha$ will denote the dimension of the irreducible representation labeled by $\alpha$, and $\rho_\alpha$ will denote the associated representation, with the fundamental property $\rho_\alpha(g_1 g_2) = \rho_\alpha(g_1)\rho_\alpha(g_2)$ for every pair of group elements $g_1, g_2 \in G$.

We remark that the group mean introduced in the main text will be explicitly unpacked here as the integral

$$\langle f(q) \rangle_{q \in G} := \int_G \mathrm{d}\mu(q) f(q), \tag{1}$$

as this helps in showing the various steps in the derivations (the integral has to be substituted with the normalized sum for the discrete degrees of freedom).

As a first step, we prove that a function $f : V \to \mathbb{C}^{d_\alpha}$ is equivariant with respect to the action of $G$ if and only if it is a fixed point of the function $\Pi_\alpha$ defined as

$$\Pi_\alpha(f, x) := \int_G \mathrm{d}\mu(q) \, \rho_\alpha(q^{-1}) f(qx), \tag{2}$$

along the $G$-orbit of $x$. Note that throughout this SI, we will usually use $g, h$ to refer to generic elements of the group $G$, while we will use $q \in G$ to refer to the dummy integration variables. More precisely, $\Pi_\alpha(f, x)$ corresponds to evaluating at the identity the orthogonal projector on $L^2(G; \mathbb{C}^{d_\alpha})$ that maps an orbit function $F_x(h) := f(hx)$ to the equivariant function $h \mapsto \rho_\alpha(h) \int_G \rho_\alpha(q^{-1}) F_x(q) \, \mathrm{d}\mu(q)$. Explicitly, the above statement is that

$$\Pi_\alpha(f, hx) = f(hx), \quad \forall h \in G \qquad \text{iff} \qquad f \text{ is equivariant along the } G\text{-orbit of } x. \tag{3}$$

Here $\alpha$ is a set of labels that represents the irreducible subspace. Let us prove this. The forward implication is immediate: if $f$ is equivariant, $f(x) = \rho_\alpha(g^{-1})f(gx)$ for all $g \in G$, then

$$\Pi_\alpha(f, x) = \int_G \mathrm{d}\mu(q) \, \rho_\alpha(q^{-1}) f(qx) = \left( \int_G \mathrm{d}\mu(q) \underbrace{\rho_\alpha(q^{-1})\rho_\alpha(q)}_{=I_{d_\alpha}} \right) f(x) = \left( \int_G \mathrm{d}\mu(q) \right) f(x) = f(x), \tag{4}$$

where $I_{d_\alpha}$ is the identity matrix in the representation space of $\alpha$, and where the last step is obtained by the normalization of the Haar integral. The other direction is done using the same strategy shown in Ref. 1. Explicitly, if $f$ is a fixed point then

$$f(hx) = \Pi_\alpha(f, hx) = \int_G \mathrm{d}\mu(q) \, \rho_\alpha(q^{-1}) f(qhx). \tag{5}$$

Now we can use the fact that $\rho_\alpha(q^{-1}) = \rho_\alpha(h)\rho_\alpha(h^{-1})\rho_\alpha(q^{-1}) = \rho_\alpha(h)\rho_\alpha((qh)^{-1})$, and the bi-invariance of the Haar integral and obtain

$$f(hx) = \rho_\alpha(h) \int_G \mathrm{d}\mu(qh)\rho_\alpha((qh)^{-1})f(qhx) = \rho_\alpha(h)f(x), \tag{6}$$

which proves that $f$ is equivariant along the $G$-orbit of $x$, thereby concluding the proof.

---

\* M. Domina, J. W. Abbott and P. Pegolo contributed equally to this work.

## A. Properties of the function $\Pi_\alpha$

We now state two important properties of $\Pi_\alpha$

- Linearity: for every pair of functions $f$ and $g$ it holds that

$$\Pi_\alpha(f+g, hx) = \Pi_\alpha(f, hx) + \Pi_\alpha(g, hx). \tag{7}$$

- Equivariance: for every $h \in G$, we have

$$\Pi_\alpha(f, x) = \rho_\alpha(h^{-1})\Pi_\alpha(f, hx). \tag{8}$$

The linearity is immediate from the definition of the function. As already shown in Ref. 1, $\Pi_\alpha$ always produces an equivariant object. This can be proven directly by using the property

$$\rho_\alpha(h^{-1})\rho_\alpha(g^{-1}) = (\rho_\alpha(h))^{-1}(\rho_\alpha(g))^{-1} = [\rho_\alpha(g)\rho_\alpha(h)]^{-1} = (\rho_\alpha(gh))^{-1} = \rho_\alpha((gh)^{-1}). \tag{9}$$

Indeed

$$\rho_\alpha(h^{-1})\Pi_\alpha(f, hx) = \int_G d\mu(q)\rho_\alpha(h^{-1})\rho_\alpha(q^{-1})f(qhx) = \int_G d\mu(qh)\rho_\alpha((qh)^{-1})f(qhx) = \Pi_\alpha(f, x), \tag{10}$$

where we used the bi-invariance of the Haar integral in $d\mu(q) = d\mu(qh)$.

## B. The equivariance error $A_\alpha$

To guarantee that $f$ is indeed a fixed point along the $G$-orbit of $x$, we can estimate the variance

$$A_\alpha(f, x) = \sqrt{\int_G \|f(qx) - \Pi_\alpha(f, qx)\|^2 d\mu(q)}, \tag{11}$$

where we use the standard $\ell^2$ norm. Before proceeding, we comment on the notation: the above can be written in terms of the standard definition of the $L^2(G; \mathbb{C}^{d_\alpha})$ norm, namely

$$A_\alpha(f, x) = \|f(\cdot x) - \Pi_\alpha(f, \cdot x)\|_{L^2(G; \mathbb{C}^{d_\alpha})}, \tag{12}$$

however we will not use this more compact notation to avoid any confusion regarding the action of the group $G$; the only exception to this choice will be for the $G$-norm of $f$, for which we will extensively use

$$\|f\|_G := \sqrt{\langle \|f\|^2 \rangle_G} = \sqrt{\int_G d\mu(q)\|f(qx)\|^2} \tag{13}$$

Trivially, $A_\alpha(f, x)$ is zero if and only if $f$ and $\Pi_\alpha(f, x)$ coincide along the $G$-orbit of $x$ *almost everywhere* (a.e.), except on sets of measure zero. Thus, from Eq. (3), we deduce that $A_\alpha(f, x) = 0$ implies that also the function $f$ is equivariant a.e.. Therefore we have that

$$A_\alpha(f, x) = 0 \qquad \text{iff} \qquad f \text{ is equivariant a.e. along the } G\text{-orbit of } x. \tag{14}$$

Moreover, if $f$ is continuous, then the a.e. condition is promoted to point-wise and we obtain

$$A_\alpha(f, x) = 0 \qquad \text{iff} \qquad f \text{ (continuous) is equivariant along the } G\text{-orbit of } x, \tag{15}$$

giving yet another definition of equivariance for continuous functions.

From a numerical perspective, however, the above expression is impractical, as it contains a double integration where terms like $f(hqx)$ arise. This would make numerical integration challenging as compositions of group elements would be required. We will prove that we can write the above equation as

$$A_\alpha(f, x) = \sqrt{\|f\|_G^2 - \|\Pi_\alpha(f, x)\|^2}, \tag{16}$$

which reduces the double integration to a single integral (implied in the definition of the function $\Pi_\alpha$) followed by an $\ell^2$ norm. In particular, the first term does not depend on the irreducible space $\alpha$ anymore, making its evaluation general and not dependent on the representation. Also, the second term is simplified with no transformation applied on the argument $x$, and each term can be evaluated by means of a single integration.

In order to prove this equation let us expand the $\ell^2$ norm in the integrand of Eq. (11) as

$$\|f(hx) - \Pi_\alpha(f, hx)\|^2 = \|f(hx)\|^2 + \|\Pi_\alpha(f, hx)\|^2 - 2\operatorname{Re}\langle f(hx) \,|\, \Pi_\alpha(f, hx)\rangle, \tag{17}$$

where we used the standard $\ell^2$ inner product. Once integrated, the first term is already the first term of Eq. (16). Therefore, we need to focus only on the remaining two. Let us continue with the second one. The first thing that we notice is that the $\ell^2$ inner product (and therefore the associated norm) is invariant with respect to the action of any orthogonal representation. Namely, this means that for any pair of vectors $v, w \in \mathbb{C}^{d_\alpha}$ we have

$$\langle v|w\rangle = v^\dagger w = v^\dagger \underbrace{\rho_\alpha^\dagger(g)\rho_\alpha(g)}_{I_{d_\alpha}} w = (\rho_\alpha(g)v)^\dagger(\rho_\alpha(g)w) = \langle \rho_\alpha(g)v \,|\, \rho_\alpha(g)w\rangle, \tag{18}$$

for every $g \in G$. Using this property and the equivariance of $\Pi_\alpha$ we can evaluate

$$\|\Pi_\alpha(f, hx)\|^2 = \|\rho_\alpha(h^{-1})\Pi_\alpha(f, hx)\|^2 = \|\Pi_\alpha(f, x)\|^2, \tag{19}$$

Now this term does not depend on the outer integration anymore and thus it is invariant under the (normalized) Haar integration over $q$:

$$\int_G \mathrm{d}\mu(q)\|\Pi_\alpha(f, x)\|^2 = \|\Pi_\alpha(f, x)\|^2 \int_G \mathrm{d}\mu(q) = \|\Pi_\alpha(f, x)\|^2. \tag{20}$$

Thus, it remains to show that

$$\int_G \mathrm{d}\mu(q)\langle f(qx) \,|\, \Pi_\alpha(f, qx)\rangle = \|\Pi_\alpha(f, x)\|^2. \tag{21}$$

This is done by using the invariance of the inner product and the equivariance of $\Pi_\alpha$ ($\Pi_\alpha(f, qx) = \rho_\alpha(q)\Pi_\alpha(f, x)$) to rewrite the left-hand side

$$\int_G \mathrm{d}\mu(q)\langle f(qx) \,|\, \Pi_\alpha(f, qx)\rangle = \int_G \mathrm{d}\mu(q)\langle \rho_\alpha(q^{-1})f(qx) \,|\, \underbrace{\rho_\alpha(q^{-1})\rho_\alpha(q)}_{I_{d_\alpha}}\Pi_\alpha(f, x)\rangle = \int_G \mathrm{d}\mu(q)\langle \rho_\alpha(q^{-1})f(qx) \,|\, \Pi_\alpha(f, x)\rangle$$

$$= \left\langle \int_G \rho_\alpha(q^{-1})f(qx)\mathrm{d}\mu(q) \,\middle|\, \Pi_\alpha(f, x)\right\rangle = \langle \Pi_\alpha(f, x) \,|\, \Pi_\alpha(f, x)\rangle = \|\Pi_\alpha(f, x)\|^2, \tag{22}$$

where we can bring the integral inside the inner product since the second term does not depend on $q$.

## C. Properties of $A_\alpha$

In this section, we comment on a few important properties of the equivariance error $A_\alpha$. First of all, the function depends only on the $G$-orbit of $x$, and not on any particular representative along that orbit. Namely,

$$A_\alpha(f, gx) = A_\alpha(f, x), \tag{23}$$

for every $g \in G$. This follows directly from its definition as

$$A_\alpha(f, gx) = \sqrt{\int_G \mathrm{d}\mu(q)\|f(qgx) - \Pi_\alpha(f, qgx)\|^2} = \sqrt{\int_G \mathrm{d}\mu(q)\|f(qx) - \Pi_\alpha(f, qx)\|^2} = A_\alpha(f, x), \tag{24}$$

obtained by first using $\mathrm{d}\mu(q) = \mathrm{d}\mu(qg)$ followed by a relabeling of the integration variable $qg \to q$. The importance of this property is that $A_\alpha(f, x)$ is independent of the particular orientation of the input $x$.

The second fundamental property is that $A_\alpha$ is invariant under the addition of any equivariant function transforming according to the irrep $\alpha$. This means that, if $f_\alpha$ is an equivariant function, then

$$A_\alpha(f + f_\alpha, x) = A_\alpha(f, x). \tag{25}$$

Again, this follows directly from the definition using the fact that the function $\Pi_\alpha$ is linear and its fixed points are exactly the equivariant functions. Therefore we have the chain of identities

$$f(hx) + f_\alpha(hx) - \Pi_\alpha(f + f_\alpha, hx) = f(hx) + f_\alpha(hx) - \Pi_\alpha(f, hx) - \underbrace{\Pi_\alpha(f_\alpha, hx)}_{=f_\alpha(hx)} = f(hx) - \Pi_\alpha(f, hx), \tag{26}$$

which, when plugged into the definition of $A_\alpha$ trivially proves the statement. This property is important because it shows that $A_\alpha(f, x)$ provides an estimate of the size of the "non-equivariant" portion of $f$, unaffected by the portion that lives in an equivariant space.

The last property is a simple bound. From the definition we have that $A_\alpha(f, x) \geq 0$, with it being 0 iff $f$ is equivariant. However, from our derived Eq. (16) we obtain the upper bound

$$0 \leq A_\alpha^2(f, x) \leq \|f\|_G^2. \tag{27}$$

This is consistent with the interpretation in Ref. 1: if we read the term $\|\Pi_\alpha(f, x)\|^2$ as the length of the equivariant part of $f(x)$ along the $G$-orbit of $x$ and $\|f\|_G$ as its full averaged length, then whenever the function is fully equivariant we will have coincidence between the two and therefore their difference will be driven to zero. On the contrary, when there is no equivariant portion, then $\|\Pi_\alpha(f, x)\|^2 = 0$ and the variance $A_\alpha(f, x)$ will show the full length $\|f\|_G$.

## II. CHARACTERS DECOMPOSITION (PETER-WEYL)

In the case in which we have a function with an arbitrary number of components $d \in \mathbb{N}^+$, $f : V \to \mathbb{C}^d$, we cannot directly apply the equivariance error defined above. However we can still extract component-wise information about membership in a given irreducible representation, as usually done by using the Peter-Weyl theorems [2]. Here we will explicitly show how this is achieved by means of the characters of the representation, $\chi_\alpha(g) = \text{Tr}(\rho_\alpha(g))$. In particular, we will use the convolution function (a flavor of the *isotypic projection* or *character projection operator* made dependent also on the input $x$) defined as

$$C_\alpha(h, f, x) := d_\alpha \int_G d\mu(q) \, \chi_\alpha(q) f(q^{-1} hx), \tag{28}$$

or, in the case of compact or discrete groups (for which unimodularity holds), equivalently

$$C_\alpha(h, f, x) := d_\alpha \int_G d\mu(q) \, \chi_\alpha(q^{-1}) f(qhx). \tag{29}$$

We will see that this function allows to build a measure that could be used element-wise on vectors of arbitrary length, and is the main ingredient for the character projection $B_\alpha$ defined in the main text.

### A. Fundamental property of the convolution function (projector with respect to the $L^2(G; \mathbb{C}^d)$ inner product)

The characterizing property of the functions $C_\alpha$ is the chain of identities

$$\int_G d\mu(q) \left\langle C_\alpha(q, f_1, x_1) \middle| C_{\alpha'}(q, f_2, x_2) \right\rangle =$$
$$= \delta_{\alpha\alpha'} \int_G d\mu(q) \left\langle f_1(qx_1) \middle| C_{\alpha'}(q, f_2, x_2) \right\rangle = \delta_{\alpha\alpha'} \int_G d\mu(q) \left\langle C_\alpha(q, f_1, x_1) \middle| f_2(qx_2) \right\rangle, \tag{30}$$

for any pair of functions $f_1, f_2 \in L^2(G; \mathbb{C}^d)$. These identities reflect the fact that the convolution function can be seen as a projector in this space, meaning that it must be self-adjoint and idempotent. We will not discuss this explicitly, but just prove these necessary identities. The prerequisite for the derivation is the Peter-Weyl orthogonality for compact groups

$$\int_G d\mu(q) \, (\rho_\alpha(q))_{mn}^* (\rho_{\alpha'}(q))_{m'n'} = \frac{\delta_{\alpha\alpha'}}{d_\alpha} \delta_{mm'} \delta_{nn'}, \tag{31}$$

which allows to straightforwardly derive also

$$\int_G \mathrm{d}\mu(q)\, \chi_\alpha^*(q)\rho_{\alpha'}(q) = \frac{\delta_{\alpha\alpha'}}{d_\alpha} I_{d_\alpha}, \qquad \text{and} \qquad \int_G \mathrm{d}\mu(q)\, \chi_\alpha^*(q)\chi_{\alpha'}(qp) = \frac{\delta_{\alpha\alpha'}}{d_\alpha}\chi_\alpha(p). \tag{32}$$

From these we can obtain the practical formula

$$\int_G \mathrm{d}\mu(q)\, \left\langle C_\alpha(q,f_1,x_1)\middle| C_{\alpha'}(q,f_2,x_2)\right\rangle = \delta_{\alpha\alpha'}d_\alpha \iint_G \mathrm{d}\mu(q_1)\mathrm{d}\mu(q_2)\, f_1^\dagger(q_1 x_1)\chi_\alpha(q_1 q_2^{-1})f_2(q_2 x_2). \tag{33}$$

Before proceeding with the derivation, we notice that this expression is extremely useful, as it reduces the number of requested integrals to only two and it allows to have the function $f_1$ and $f_2$ evaluated on only one grid for the transformations $q$, where the composition $q_1 q_2^{-1}$ is incorporated in the character only. Without this formula, the evaluation of the left-hand side of Eq. (30) would require a triple integration, and the functions $f_1$ and $f_2$ would be evaluated on a grid of composed transformations, making numerical integration more challenging. For these reasons, not only this formula is used to derive Eq. (30), but it will also be used in numerical experiments. Proceeding now with the derivation, we first write $C_\alpha$ as

$$C_\alpha(h,f,x) = d_\alpha \int_G \mathrm{d}\mu(q)\, \chi_\alpha(hq^{-1})f(qx), \tag{34}$$

where we changed variables as $g^{-1}h = q$, which implies $g = hq^{-1}$, and we used the bi-invariance of the Haar measure. Therefore we can expand the left-hand side of Eq. (33)

$$\int_G \mathrm{d}\mu(q)\, \left\langle C_\alpha(q,f_1,x_1)\middle| C_{\alpha'}(q,f_2,x_2)\right\rangle = d_\alpha d_{\alpha'} \iint_G \mathrm{d}\mu(q_1)\mathrm{d}\mu(q_2)\, f_1^\dagger(q_1 x_1)f_2(q_2 x_2) \int_G \mathrm{d}\mu(h)\, \chi_\alpha(hq_1^{-1})\chi_\alpha(hq_2^{-1}). \tag{35}$$

Now by defining the new variable $q' = qq_1^{-1}$, which implies $qq_2^{-1} = q'q_1 q_2^{-1}$, and by using the bi-invariance of the Haar measure, $\mathrm{d}\mu(q') = \mathrm{d}\mu(q)$ the inner integration can be carried out by means of the second of Eqs. (32), leading directly to our target Eq. (33).

From Eq. (33), the chain of Eq. (30) can be straightforwardly derived by simple change of variable. For example, to obtain the last equality, we can simply define $q' = (q_1 q_2^{-1})^{-1}$, and recalling that $\chi_\alpha(q^{-1}) = \chi_\alpha^*(q)$ get to

$$\int_G \mathrm{d}\mu(q)\, \left\langle C_\alpha(q,f_1,x_1)\middle| C_{\alpha'}(q,f_2,x_2)\right\rangle = \delta_{\alpha\alpha'}d_\alpha \int_G \mathrm{d}\mu(q_2)\left(\underbrace{\int_G \mathrm{d}\mu(q)\, \chi_\alpha(q)f_1(q^{-1}q_2 x_1)}_{=C_\alpha(q_2,f_1,x_1)}\right)^\dagger f_2(q_2 x_2), \tag{36}$$

which is exactly the desired outcome for $q_2 = q$. Similarly the first equivalence in Eq. (30) is obtained again from Eq. (16) by means of the change of variable $q' = q_1 q_2^{-1}$.

## B. Decomposing the function $f$

In this section we will show how the function $C_\alpha$ can be used to decompose a function $f : V \to \mathbb{C}^d$. Our starting point is that the function $f$ is square-integrable on the full orbit $G$-orbit of $x$, and thus admits a decomposition

$$f(hx) = \sum_\alpha f_\alpha(hx), \tag{37}$$

for any $h \in G$. Here, $f_\alpha(hx)$ are the *isotypic* channels of $f$. Each of the components of $f_\alpha(hx)$ can be expanded as a linear combination of equivariant functions as

$$(f_\alpha(hx))_k = \sum_{n\in\mathbb{N}} a_{kn\alpha}^T \cdot \phi_{kn\alpha}(hx). \tag{38}$$

where the function $\phi_{kn\alpha}(gx) \in \mathbb{C}^{d_\alpha}$ are equivariant functions for the irreducible representation labeled by $\alpha$, namely it holds that $\phi_{kn\alpha}(gx) = \rho_\alpha(g)\phi_{kn\alpha}(x)$, for every $g \in G$. The index $n \in \mathbb{N}$ (or a countable index set) enumerates a complete countable basis of the corresponding $L^2$ isotypic subspace. Here $a_{kn\alpha} \in \mathbb{C}^{d_\alpha}$ are arbitrary vectors, and can be thought as basis expansion coefficients. With such $f$ we will now prove that

$$C_\alpha(h,f,x) = f_\alpha(hx), \tag{39}$$

namely the convolution function extracts the isotypic components. The proof is quite straightforward and it directly uses the decomposition of $f$ and the definition of $C_\alpha$. Indeed we have

$$C_\alpha(h, f, x) = d_\alpha \sum_{\alpha' n} a_{kn\alpha'}^T \cdot \left[ \left( \underbrace{\int_G \mathrm{d}\mu(q)\, \chi_\alpha(hq^{-1}) \rho_{\alpha'}(q)}_{= \delta_{\alpha\alpha'} \rho_\alpha(h)/d_\alpha} \right) \phi_{kn\alpha'}(x) \right] = \sum_n a_{kn\alpha}^T \cdot \underbrace{\rho_\alpha(h) \phi_{kn\alpha}(x)}_{\phi_{kn\alpha}(hx)} = f_\alpha(hx), \qquad (40)$$

where in the first step we used $\mathrm{d}\mu(q) = \mathrm{d}\mu(qh^{-1})$, followed by $\chi_\alpha(hq^{-1}) = \chi_\alpha^*(qh^{-1})$ and $\rho_\alpha(q) = \rho_\alpha(qh^{-1})\rho_\alpha(h)$, and the application Eqs. (32) to evaluate the integral. Instead, in the second step we simply used the equivariance of $\phi_{kn\alpha}$.

Now, we can use Eq. (39) and the relations of Eq. (30) to derive the relation

$$\|f\|_G^2 = \sum_\alpha \int_G \mathrm{d}\mu(q) \, \|f_\alpha(qx)\|^2 = \sum_\alpha \int_G \mathrm{d}\mu(q) \, \|C_\alpha(q, f, x)\|^2. \qquad (41)$$

Again, this is proved straightforwardly by using the explicit decomposition of the function $f$:

$$\|f\|_G^2 = \sum_{\alpha\alpha'} \int_G \mathrm{d}\mu(q) \, \langle C_\alpha(q, f, x) \,|\, C_{\alpha'}(q, f, x) \rangle = \sum_\alpha \int_G \mathrm{d}\mu(q) \, \|C_\alpha(q, f, x)\|^2, \qquad (42)$$

where, we obtained the last step by applying twice the identities in Eq. (30). We therefore define the character projections as

$$B_\alpha(f, x) := \int_G \|C_\alpha(q, f, x)\|^2 \mathrm{d}\mu(q) = d_\alpha^2 \left\langle \left\| \left\langle \chi_\alpha(q'^{-1}) t(q'qx) \right\rangle_{q' \in G} \right\|_2^2 \right\rangle_{q \in G}. \qquad (43)$$

where the right-hand side, in terms of averages, links this expression with Eq. (4) of the main text. Using Eq. (33) we obtain the practical formula

$$B_\alpha(f, x) = d_\alpha \iint_G \mathrm{d}\mu(q_1)\mathrm{d}\mu(q_2) \, f^\dagger(q_1 x) \chi_\alpha(q_1 q_2^{-1}) f(q_2 x), \qquad (44)$$

On the one hand, we have that

$$\|f\|_G^2 = \sum_\alpha B_\alpha(f, x). \qquad (45)$$

On the other hand, we can define the isotypic residual (analogous to the equivariance error $A_\alpha$ but for the character projections) as

$$R_\alpha(f, x) := \sqrt{\|f\|_G^2 - B_\alpha(f, x)}, \qquad (46)$$

which is equal to zero if, and only if, $f(gx) = f_\alpha(gx)$ a.e., namely the function $f$ is constructed from a combination of terms that belong to the same irreducible representation along the full $G$-orbit of $x$, but for $\mu$-null sets. Indeed, from Eqs. (39) and (42), if $f(gx) = f_\alpha(gx)$ then $R_\alpha(f, x)$ is trivially zero. On the contrary, if $R_\alpha(f, x) = 0$ then, from Eq. (42) we have

$$R_\alpha(f, x) = \sqrt{\|f\|_G^2 - B_\alpha(f, x)} = \sqrt{\sum_{\alpha'} B_{\alpha'}(f, x) - B_\alpha(f, x)} = \sqrt{\sum_{\substack{\alpha' \\ \alpha' \neq \alpha}} B_{\alpha'}(f, x)} = 0. \qquad (47)$$

Since all the integrands in the last steps are non-negative and made by norms, their average can be zero only if $C_{\alpha'}(h, f, x) = 0$ a.e., for every $\alpha' \neq \alpha$. Again, if $f$ is continuous, then also $C_{\alpha'}(h, f, x)$ are continuous and the a.e. condition is promoted to point-wise (because also the characters are continuous and the convolution preserves continuity). Therefore we have that

$$R_\alpha(f, x) = 0 \qquad \text{iff} \qquad f \text{ lies in the } \alpha\text{-sector a.e. along the } G\text{-orbit of } x, \qquad (48)$$

and if $f$ is continuous then the a.e. condition is promoted to point-wise and we obtain

$$R_\alpha(f, x) = 0 \qquad \text{iff} \qquad f \text{ (continuous) lies in the } \alpha\text{-sector along the full } G\text{-orbit of } x. \qquad (49)$$

Note that in this case $f$ can be a function of arbitrary dimensionality, i.e. it could also be a portion of a larger vector or even just one component. As an immediate consequence, $f$ lies in the $\alpha$ irreducible sector iff $\|f\|_G^2 = B_\alpha(f, x)$.

### III. THE CASE $d = d_\alpha$

In the case in which the function $f$ is assumed to be equivariant with respect to the $\alpha$ irreducible representation we can compare Eqs. (16) and (46) as $d = d_\alpha$. In particular, we can see how the term $\|f\|_G^2$ is in common between the two. In this special case we can observe how

$$0 \le R_\alpha^2(f, x) \le A_\alpha^2(f, x) \le \|f\|_G^2. \tag{50}$$

This inequality can be intuitively read from the fact that a necessary condition for the equivariance of $f$ is that it fully lies on the $\alpha$-sector. Indeed, if $f$ is equivariant, then applying $C_\alpha$ leads to

$$C_\alpha(h, f, x) = d_\alpha \int_G d\mu(q) \, \chi_\alpha(q) f(q^{-1}hx) = \underbrace{d_\alpha \int_G d\mu(q) \, \chi_\alpha^*(q^{-1}) \rho_\alpha(q)}_{= I_{d_\alpha}} f(hx) = f(hx), \tag{51}$$

namely the $\alpha$-isotypic component of $f$ coincides with $f$ itself. Therefore, whenever $A_\alpha(f, x)$ vanishes, so does $R_\alpha(f, x)$. Now we can prove the inequality above by just proving

$$\|\Pi_\alpha(f, x)\|^2 \le B_\alpha(f, x), \tag{52}$$

which implies that the equivariant component of $f$ is always at most its full isotypic component. The proof is straightforward by noticing that the evaluation of $\Pi_\alpha(f, x)$ over $f$ is equivalent to the evaluation of $\Pi_\alpha$ over the $\alpha$-isotypic component of $f$, which is exactly $C_\alpha(h, f, x)$, namely

$$\Pi_\alpha(f, x) = \int_G d\mu(q) \, \rho_\alpha(q^{-1}) f(qx) = \int_G d\mu(q) \, \rho_\alpha(q^{-1}) C_\alpha(q, f, x). \tag{53}$$

Intuitively, being in the $\alpha$-sector is a necessary condition for being equivariant, then the projection of $f$ on the $\alpha$-sector (which is exactly $C_\alpha(h, f, x)$) must be at least as large as the projection of $f$ on the equivariant subspace (which is exactly $\Pi_\alpha(f, x)$). This expression is easily obtained by using the definition of $C_\alpha$:

$$\begin{aligned}
\int_G d\mu(q) \, \rho_\alpha(q^{-1}) C_\alpha(q, f, x) &= d_\alpha \int_G d\mu(q) \, \rho_\alpha(q^{-1}) \int_G d\mu(q') \, \chi_\alpha(qq'^{-1}) f(q'x) \\
&= d_\alpha \underbrace{\int_G d\mu(q') \, \rho_\alpha(q'^{-1}) f(q'x)}_{= \Pi_\alpha(f, x)} \underbrace{\int_G d\mu(q'q^{-1}) \, \chi_\alpha^*(q'q^{-1}) \rho_\alpha(q'q^{-1})}_{= 1/d_\alpha} = \Pi_\alpha(f, x),
\end{aligned} \tag{54}$$

obtained by using $\rho_\alpha(q) = \rho_\alpha(q'^{-1}) \rho_\alpha(q'q^{-1})$, the bi-invariance and unimodularity $d\mu(q) = d\mu(q'q^{-1})$ and the properties of the characters in Eqs. (32). The inequality is then obtained by using the Cauchy-Schwarz/Jensen inequality (for the normalized Haar measure) as

$$\|\Pi_\alpha(f, x)\|^2 = \left\| \int_G d\mu(q) \, \rho_\alpha(q^{-1}) C_\alpha(q, f, x) \right\|^2 \le \int_G d\mu(q) \, \|\rho_\alpha(q^{-1}) C_\alpha(q, f, x)\|^2 = \int_G d\mu(q) \, \|C_\alpha(q, f, x)\|^2 = B_\alpha(f, x), \tag{55}$$

where we used the invariance of the $l^2$ norm with respect to the action of any orthogonal representation to get the last step.

### A. Case $d = d_\alpha = 1$

In this sub-case of the previous one, we have that the function $f$ is a scalar function and the representation $\alpha$ is one-dimensional. In this case, the equivariant condition and the character table really coincide, because $\rho_\alpha(g) = \chi_\alpha(g)$, and therefore the character inherits the composition property of the representation. Thus, we can use $\chi_\alpha(q_1 q_2^{-1}) = \rho_\alpha(q_1) \rho_\alpha(q_2^{-1})$ in Eq. (33) to get

$$B_\alpha(f, x) = \iint_G d\mu(q_1) d\mu(q_2) \, f^*(q_1 x) \rho_\alpha(q_1) \rho_\alpha(q_2^{-1}) f(q_2 x) = \left| \int_G d\mu(q) \, \rho_\alpha(q) f(qx) \right|^2 = \|\Pi_\alpha(f, x)\|^2. \tag{56}$$

This also implies that the equivariance error and the isotypic residual coincide.

## IV.  THE CASE OF THE ORTHOGONAL GROUP $O(3)$

In this section we apply the previous results to the orthogonal group $O(3)$, which is the relevant symmetry group for three-dimensional point clouds. We parametrize elements of $O(3)$ using $ZYZ$ Euler angles $\alpha, \gamma \in [0, 2\pi)$ and $\beta \in [0, \pi]$, together with a discrete label $s = \pm 1$ indicating whether the transformation is proper ($s = 1$) or improper ($s = -1$, including an inversion). A general roto-inversion $q \in O(3)$ is therefore written as $q = \Phi(R, s) \equiv \Phi(R(\alpha, \beta, \gamma), s)$, where $R \in SO(3)$ is the proper rotation associated with the Euler angles $\alpha, \beta, \gamma$. Here $\Phi(R, s)$ denotes the fixed parametrization of $O(3)$ in terms of the subgroup $SO(3)$ and the discrete label $s$.

Irreducible representations of $O(3)$ are labeled by $\alpha = (\lambda, \sigma)$, where $\lambda \in \mathbb{N}$ denotes the angular momentum (spherical-harmonic degree) and $\sigma = \pm 1$ represents the parity, proper ($\sigma = +1$) or pseudo ($\sigma = -1$). With this convention, group elements are represented by the Wigner $D^\lambda \in \mathbb{R}^{(2\lambda+1) \times (2\lambda+1)}$ matrices (we use their real representation), namely

$$\rho_\alpha(q) = \rho_{\lambda\sigma}(\Phi(R, s)) = \left(\sigma(-1)^\lambda\right)^{\frac{1-s}{2}} D^\lambda(R). \tag{57}$$

The dimension of the irreducible representation labeled by $\alpha = (\lambda, \sigma)$ is therefore $d_\alpha = 2\lambda + 1$.

The Haar measure on $O(3)$ can be expressed in terms of the Haar measure on the subgroup $SO(3)$. For a function $f$ defined over $O(3)$, it holds that

$$\int_{O(3)} f(q) \mathrm{d}\mu(q) = \frac{1}{2} \sum_{s=\pm 1} \int_{SO(3)} f(\Phi(R, s)) \mathrm{d}\mu_{SO(3)}(R) = \frac{1}{2} \int_{SO(3)} \left[ \sum_{s=\pm 1} f(\Phi(R, s)) \right] \mathrm{d}\mu_{SO(3)}(R). \tag{58}$$

In terms of the Euler angles, the Haar measure on $SO(3)$ can be written as

$$\mathrm{d}\mu_{SO(3)}(R(\alpha, \beta, \gamma)) = \frac{1}{8\pi^2} \mathrm{d}\alpha \, \mathrm{d}(\cos \beta) \, \mathrm{d}\gamma, \qquad \int_{SO(3)} \equiv \int_0^{2\pi} \mathrm{d}\alpha \int_{-1}^1 \mathrm{d}(\cos \beta) \int_0^{2\pi} \mathrm{d}\gamma. \tag{59}$$

This expression provides a practical way to evaluate integrals over $O(3)$, such as those defining $\Pi_\alpha(f, x)$ and $\|f\|_G^2$, and thus also $A_\alpha(f, x)$.

### A.  Computation of $B_\alpha$ for $O(3)$

To evaluate the character projection $B_\alpha$ for $O(3)$, we need to compute the characters of the irreducible representations. For $O(3)$, the characters can be expressed in terms of the Wigner $D$-matrices as

$$\chi^{\lambda\sigma}(q) = \mathrm{Tr}(\rho_{\lambda\sigma}(\Phi(R, s))) = \left(\sigma(-1)^\lambda\right)^{\frac{1-s}{2}} \mathrm{Tr}(D^\lambda(R)) = \left(\sigma(-1)^\lambda\right)^{\frac{1-s}{2}} \chi^\lambda_{SO(3)}(R). \tag{60}$$

The last term $\chi^\lambda_{SO(3)}(R(\alpha, \beta, \gamma))$ is the character of the irreducible representation of $SO(3)$ evaluated on the proper rotation $R(\alpha, \beta, \gamma) \in SO(3)$ and labeled by $\lambda$. These characters depend on a single angle [3] (the angle of rotation $\omega$) as

$$\chi^\lambda_{SO(3)}(R(\alpha, \beta, \gamma)) = \frac{\sin\left((2\lambda+1)\frac{\omega}{2}\right)}{\sin\frac{\omega}{2}}, \tag{61}$$

where $\omega$ is written in terms of the Euler angles as

$$\cos\frac{\omega}{2} = \cos\frac{\beta}{2} \cos\frac{\alpha+\gamma}{2}. \tag{62}$$

With these definitions, we can now evaluate the character projection $B_\alpha$ for $O(3)$ by using Eq. (44). We have

$$B_{\lambda\sigma}(f, x) = \int_{O(3)} \|C_{\lambda\sigma}(h, f, x)\|^2 \mathrm{d}\mu(h) = (2\lambda+1) \iint_{O(3)} \mathrm{d}\mu(q_1) \mathrm{d}\mu(q_2) \, f^\dagger(q_1 x) \chi^{\lambda\sigma}(q_1 q_2^{-1}) f(q_2 x). \tag{63}$$

Using the defining property of the representations, we can write

$$\chi^{\lambda\sigma}(q_1 q_2^{-1}) = \mathrm{Tr}(\rho_{\lambda\sigma}(q_1 q_2^{-1})) = \mathrm{Tr}(\rho_{\lambda\sigma}(q_1) \rho_{\lambda\sigma}(q_2)^{-1}) = \mathrm{Tr}(\rho_{\lambda\sigma}(q_2) \rho_{\lambda\sigma}(q_1)^{-1}) = \chi^{\lambda\sigma}(q_2 q_1^{-1}), \tag{64}$$

namely the characters are class functions. By using this property, together with Eqs. (60), (58) and a simple relabeling of the integration variables, we obtain the final formula for the character projection $B_\alpha$ for $O(3)$:

$$
\begin{aligned}
B_{\lambda\sigma}(f,x) = \frac{(2\lambda+1)}{4} \iint\limits_{SO(3)} \chi^\lambda_{SO(3)}(R_1 R_2^{-1}) \Bigg[ & \sum_{s=\pm 1} f^\dagger(\Phi(R_1,s)x) f(\Phi(R_2,s)x) \\
& + 2\sigma(-1)^\lambda f^\dagger(\Phi(R_1,+1)x) f(\Phi(R_2,-1)x) \Bigg] \mathrm{d}\mu_{SO(3)}(R_1)\mathrm{d}\mu_{SO(3)}(R_2),
\end{aligned}
\tag{65}
$$

written in terms of two nested integrals over $SO(3)$. Note that the only term containing the composition of the two transformations $R_1$ and $R_2$ is the character $\chi^\lambda_{SO(3)}(R_1 R_2^{-1})$, while the function $f$ is evaluated on the individual transformations $R_1$ and $R_2$. This follows directly from the class-function property of the characters, and it is crucial for the numerical evaluation of this quantity.

## V. INTEGRATION SCHEME FOR THE INTEGRALS OVER $O(3)$

We have explored how the integrals over $O(3)$ can be reduced to integrals over $SO(3)$, parametrized by the Euler angles. We now discuss how to make this numerically tractable. Our main assumption will be that the integrand function is smooth enough so that it is effectively band-limited in the Fourier space of $SO(3)$, meaning that it can be well approximated by a finite number of Wigner $D$-matrices (by the Peter-Weyl theorem) with a small residual that can be made arbitrarily small by increasing the number of terms in the expansion. We will also not write the dependence on the input $x$ for simplicity, as it is not relevant for the discussion. Under these assumptions we have, for fixed $s$,

$$
f(\Phi(R,s)) \approx \sum_{\lambda=0}^L \sum_{m,n=-\lambda}^\lambda \hat{f}_{mn\lambda s} D^\lambda_{mn}(R).
\tag{66}
$$

For the theoretical analysis we will use the standard complex representation of the Wigner $D$-matrices, as the integration scheme is more straightforwardly defined in this case. However, since the real and complex representations are related by a simple change of basis that keeps the orthogonality properties of the Wigner $D$-matrices, the same integration scheme can be also used for the real representation.

Integrating a function like $f$ over $SO(3)$ requires proper handling of the dependence on the Euler angles. In complex notation we have that the Wigner $D$-matrices can be written as

$$
D^\lambda_{mn}(\alpha,\beta,\gamma) = e^{-im\alpha} d^\lambda_{mn}(\beta) e^{-in\gamma}, \qquad \text{with} \quad |m|, |n| \leq \lambda,
\tag{67}
$$

where $d^\lambda_{mn}(\beta)$ are the Wigner small $d$-matrices, which are real-valued functions of the angle $\beta$. The dependence on $\gamma$ appears as a simple Fourier mode $e^{-in\gamma}$, while the dependence on $\alpha$ and $\beta$ can be viewed as a joint dependence over the polar coordinates of the sphere.

Assuming that the effective bandwidth of the function $f$ is $L$, we can thus adopt a product quadrature scheme:

- For the polar angles $(\alpha, \beta)$, we can use the Lebedev quadrature [4], which is designed to efficiently integrate spherical harmonics up to a certain degree $L_{\text{Leb}}$.

- For the angle $\gamma$, we can use the trapezoidal rule, which integrates exactly Fourier modes $e^{-in\gamma}$ up to frequency $n = L$, using $L_{\text{trap}} = L + 1$ number of points.

One could think that applying this direct quadrature scheme could generally fail for the integration of general functions, as the Lebedev quadrature is only exact for scalar spherical harmonics. The reason is that, for $n \neq 0$, the factor $e^{-im\alpha} d^\lambda_{mn}(\beta)$ does not belong to the finite-dimensional span of ordinary (spin-0) spherical harmonics of degree $\leq \lambda$. More precisely, the $(\alpha, \beta)$-dependence of a Wigner $D$-matrix element can be identified (up to a convention-dependent phase and normalization) with a spin-weighted spherical harmonic. Using the standard relation

$$
{}_{-n}Y_{\lambda m}(\beta,\alpha) = (-1)^n \sqrt{\frac{2\lambda+1}{4\pi}} D^\lambda_{m,n}(\alpha,\beta,0) = (-1)^n \sqrt{\frac{2\lambda+1}{4\pi}} e^{-im\alpha} d^\lambda_{m,n}(\beta),
\tag{68}
$$

shows that the Lebedev quadrature would be exact only for the case $n = 0$, which corresponds to the case of ordinary spherical harmonics. However, we will prove that this is not a limitation and that a Lebedev and trapezoidal

quadrature can be used together to integrate products of Wigner $D$-matrices exactly. Indeed, we can prove that this scheme integrate exactly the condition

$$\int_{SO(3)} D_{m_1 n_1}^{\lambda_1 *}(R) D_{m_2 n_2}^{\lambda_2}(R) \mathrm{d}\mu_{SO(3)}(R) = \frac{1}{2\lambda_1 + 1} \delta_{\lambda_1 \lambda_2} \delta_{m_1 m_2} \delta_{n_1 n_2}, \tag{69}$$

of orthogonality of the Wigner $D$-matrices. This will also imply that we can exactly integrate one Wigner D matrix only, as it can be seen as a special case of the above by setting $\lambda_2 = m_2 = n_2 = 0$ and using $D_{00}^0(R) = 1$. We remark that a related problem has been investigated in Ref. 5: here we will follow a parallel (yet equivalent) derivation tailored to the specific case of quadratic products of Wigner $D$-matrices, but we mention that the result is essentially the same. We will show now that the following statement holds.

*The product quadrature scheme defined by the Lebedev quadrature for the angles $\alpha, \beta$ and the trapezoidal rule for the angle $\gamma$ is exact for the integration of products of Wigner $D$-matrices up to bandwidth $L$ if the Lebedev quadrature is exact for the integration of ordinary spherical harmonics up to degree $2L$ and the trapezoidal rule uses at least $2L+1$ points.*

To show this, we can first decompose the product of two Wigner $D$-matrices as

$$D_{m_1 n_1}^{\lambda_1}(\alpha, \beta, \gamma) D_{m_2 n_2}^{\lambda_2 *}(\alpha, \beta, \gamma) = e^{-i(m_1 - m_2)\alpha} e^{i(n_1 - n_2)\gamma} d_{m_1 n_1}^{\lambda_1}(\beta) d_{m_2 n_2}^{\lambda_2}(\beta). \tag{70}$$

This expression shows that performing the trapezoidal rule for the angle $\gamma$ with at least $2L+1$ points is sufficient to guarantee the exact integration of the Fourier mode $e^{i(n_1 - n_2)\gamma}$, so that we obtain

$$\int_0^{2\pi} e^{i(n_1 - n_2)\gamma} \mathrm{d}\gamma = 2\pi \delta_{n_1 n_2}. \tag{71}$$

Since this is numerically exact (up to machine precision) we are left with an integration of the form

$$\int_0^{2\pi} \mathrm{d}\alpha \int_{-1}^1 \mathrm{d}(\cos\beta) \; e^{-i(m_1 - m_2)\alpha} d_{m_1 n}^{\lambda_1}(\beta) d_{m_2 n}^{\lambda_2}(\beta) = \int_0^{2\pi} \mathrm{d}\alpha \int_{-1}^1 \mathrm{d}(\cos\beta) \; \left( e^{-i m_1 \alpha} d_{m_1 n}^{\lambda_1}(\beta) \right) \left( e^{i m_2 \alpha} d_{m_2 n}^{\lambda_2}(\beta) \right). \tag{72}$$

We can see that the integrand is proportional to the product of two spin-weighted spherical harmonics, as it holds that

$$e^{i m_2 \alpha} d_{m_2 n}^{\lambda_2}(\beta) = \left( e^{-i m_2 \alpha} d_{m_2 n}^{\lambda_2}(\beta) \right)^* = {}_{-n} Y_{\lambda_2 m_2}^*(\beta, \alpha) = (-1)^{n + m_2} {}_n Y_{\lambda_2, -m_2}(\beta, \alpha), \tag{73}$$

namely

$$\left( e^{-i m_1 \alpha} d_{m_1 n}^{\lambda_1}(\beta) \right) \left( e^{i m_2 \alpha} d_{m_2 n}^{\lambda_2}(\beta) \right) \propto {}_{-n} Y_{\lambda_1 m_1}(\beta, \alpha) \; {}_n Y_{\lambda_2, -m_2}(\beta, \alpha). \tag{74}$$

for some unessential normalization and phase factor. Now we can use the Clebsch-Gordan (CG) decomposition of the product of two spin-weighted spherical harmonics, which follows from the relation between spin-weighted harmonics and Wigner $D$-matrices [6] together with the standard CG series for products of Wigner $D$-matrices [3]

$$_{s_1} Y_{\lambda_1 m_1}(\beta, \alpha) \; {}_{s_2} Y_{\lambda_2 m_2}(\beta, \alpha) = \sum_{\lambda = |\lambda_1 - \lambda_2|}^{\lambda_1 + \lambda_2} \sum_{m = -\lambda}^{\lambda} C_{\lambda_1 m_1, \lambda_2 m_2}^{\lambda m} C_{\lambda_1 s_1 \lambda_2 s_2}^{\lambda, s_1 + s_2} \; {}_{s_1 + s_2} Y_{\lambda m}(\beta, \alpha), \tag{75}$$

where $C_{\lambda_1 m_1, \lambda_2 m_2}^{\lambda m}$ and $C_{\lambda_1 s_1 \lambda_2 s_2}^{\lambda, s_1 + s_2}$ are the CG coefficients. Applying this decomposition to our case, and because $-n + n = 0$, the product of the two spin-weighted spherical harmonics above can be decomposed as a linear combination of ordinary spherical harmonics. Explicitly

$$\left( e^{-i m_1 \alpha} d_{m_1 n}^{\lambda_1}(\beta) \right) \left( e^{i m_2 \alpha} d_{m_2 n}^{\lambda_2}(\beta) \right) \propto \sum_{l = |\lambda_1 - \lambda_2|}^{\lambda_1 + \lambda_2} \sum_{m = -l}^{l} c_{lm} Y_{lm}(\beta, \alpha), \tag{76}$$

for some coefficient $c_{lm}$ that depend on the CG coefficients. Therefore, we are left with the integration of ordinary spherical harmonics, which is guaranteed to be integrated exactly by a Lebedev quadrature of the appropriate order. If the maximum degree of the Wigner $D$-matrices in the integrand is $L$, then the maximum degree of the ordinary

spherical harmonics in the CG decomposition is $2L$, so we need a Lebedev quadrature that is exact for spherical harmonics up to degree $2L$, which concludes the proof. $\qquad\square$

The above statement provides justification and theoretical foundation to use the computationally efficient Lebedev quadrature to perform the integrals over $O(3)$. Therefore, we define a quadrature parameter $L_{\text{quad}}$ that controls the order of the Lebedev grid and the number of points in the trapezoidal rule, such that the quadrature scheme is exact for the integration of products of Wigner $D$-matrices up to bandwidth $L_{\text{quad}}$. Practically, this means that we choose the smallest Lebedev quadrature of order $L_{\text{Leb}}$ such that $L_{\text{Leb}} \geq 2L_{\text{quad}}$, and we choose the number of points in the trapezoidal rule $L_{\text{trap}}$ such that $L_{\text{trap}} = 2L_{\text{quad}} + 1$.

In particular, to compute $A_{\lambda\sigma}(f, x)$, it is reasonable to assume that the effective bandwidth of the function is $L_f \geq \lambda$ (since it is a test of equivariance) and use it as the quadrature parameter $L_{\text{quad}} = L_f$. Conversely, to evaluate $B_{\lambda\sigma}(f, x)$ we need to take into account the maximum order, $L_\chi$, of the character that we want to expand. Since it is reasonable to assume that we want a decomposition that *exceeds* the bandwidth of the function, we set $L_{\text{quad}} = L_\chi$.

## VI.   CLOSED FORM FOR THE SYMMETRY PURIFICATION OF THE READOUT

Following the discussion of Sec. III.C of the main text, we derive the purification protocol and the closed-form expression for the readout weights after purification. We begin by rewriting the equivariance error in a form that exposes its quadratic structure and is suitable for averaging over the group. In particular, we prove that the equivariance error can be written as

$$A_\alpha^2(f, x) = \int_G \mathrm{d}\mu(q) \left\| \rho_\alpha(q^{-1}) f(qx) - \int_G \mathrm{d}\mu(q') \, \rho_\alpha(q'^{-1}) f(q'x) \right\|^2. \tag{77}$$

This can be shown starting from the definition of $A_\alpha$ in Eq. (11), by noting that the $\ell^2$ norm is invariant under the action of any orthogonal representation, and by using the bi-invariance of the Haar measure:

$$\begin{aligned}
A_\alpha^2(f, x) &= \int_G \mathrm{d}\mu(q) \left\| f(qx) - \int_G \mathrm{d}\mu(q') \, \rho_\alpha(q'^{-1}) f(q'qx) \right\|^2 \\
&= \int_G \mathrm{d}\mu(q) \left\| \rho_\alpha(q^{-1}) f(qx) - \int_G \underbrace{\mathrm{d}\mu(q')}_{=\mathrm{d}\mu(q'q)} \, \underbrace{\rho_\alpha(q^{-1})\rho_\alpha(q'^{-1})}_{=\rho_\alpha((q'q)^{-1})} f(q'qx) \right\|^2,
\end{aligned} \tag{78}$$

which leads to the desired expression by a simple relabeling of the integration variable $q'q \to q'$. We will use this expression directly as the second term of the purification loss. In general, the targets for the sample $i$ can be written as a concatenation of different equivariant channels, each of them corresponding to a different irreducible of the group. We can write

$$y_i = (y_i^{(1)}, \ldots, y_i^{(K)}) \qquad \text{with} \quad y_i^{(k)} \in \mathbb{R}^{d_k}, \tag{79}$$

such that, under a group transformation $g \in G$, the target transforms as $y_i \xrightarrow{g} \rho(g)y_i = (\rho_1(g)y_i^{(1)}, \ldots, \rho_K(g)y_i^{(K)})$, where $\rho_k \in \mathbb{R}^{d_k \times d_k}$ is the irreducible representation of the $k$-th channel, and where $\rho(g) = \bigoplus_{k=1}^K \rho_k(g)$ is the block-diagonal representation of the whole target. The model produces a prediction $\hat{y}(x_i)$ that has the same structure as the target, and we can write $\hat{y}(x_i) = (\hat{y}^{(1)}(x_i), \ldots, \hat{y}^{(K)}(x_i))$, obtained by a block diagonal readout linear layer $\theta$ applied to a general featurized $\phi(x)$, namely

$$\hat{y}(x_i) = \theta^T \phi(x_i) = (\theta_1^T \phi(x_i), \ldots, \theta_K^T \phi(x_i)), \qquad \text{with} \quad \theta_k \in \mathbb{R}^{(d_{\text{model}}+1) \times d_k} \quad \text{and} \quad \phi(x_i) \in \mathbb{R}^{(d_{\text{model}}+1)}. \tag{80}$$

where we also included the possibility of a bias term in the readout layer, such that $\theta_k^T \phi(x_i) = \widetilde{\theta}_k^T \widetilde{\phi}(x_i) + b_k$, with $\widetilde{\theta}_k$ and $\widetilde{\phi}(x_i)$ being the non-bias part of the readout layer and of the featurized input, respectively. We remark that this is a sensible because $\phi(x_i)$ is not only not necessarily equivariant, but it can also contain terms that are constant and do not depend on the input $x_i$.

Now, following the procedure of the main text, the general purification loss can be written as

$$L = L_\mu + \gamma L_\sigma, \tag{81}$$

with

$$L_\mu =: \sum_{k=1}^{K} \eta_k^{(\mu)} L_k^{(\mu)}, \qquad \text{with} \quad L_k^{(\mu)} := \sum_i u_{i,k}^{(\mu)} \int_G \mathrm{d}\mu(q) \left\| \rho_k(q^{-1}) \hat{y}^{(k)}(qx_i) - y_i^{(k)} \right\|^2. \tag{82}$$

Here $\eta_k^{(\mu)}$ is a weighting factor for the different channels of the target, and with $u_{i,k}^{(\mu)}$ being the standard weighting factor for the different samples for each channel $k$. This term is the one that enforces that the backtransformed prediction $\rho_k(q^{-1}) \hat{y}^{(k)}(qx_i)$ over the transformed input is anchored to the target $y_i^{(k)}$ for all the transformations $q \in G$. The second term of the loss is defined analogously as

$$L_\sigma =: \sum_{k=1}^{K} \eta_k^{(\sigma)} L_k^{(\sigma)}, \qquad \text{with} \quad L_k^{(\sigma)} := \sum_i u_{i,k}^{(\sigma)} \int_G \mathrm{d}\mu(q) \left\| \rho_k(q^{-1}) \hat{y}^{(k)}(qx_i) - \int_G \mathrm{d}\mu(q') \, \rho_k(q'^{-1}) \hat{y}^{(k)}(q'x_i) \right\|^2, \tag{83}$$

which is essentially Eq. (77) applied to the prediction $\hat{y}^{(k)}$, and as such it enforces the equivariance of the prediction itself. The factor $\gamma$ controls the relative importance of the two terms of the loss.

In order to find a closed form for the optimal readout layer $\theta$ after purification, we can first notice that losses for different channels $k$ are decoupled (i.e., all the terms are block diagonal and independent) and therefore we can focus on optimizing the single loss term $L_k := L_k^{(\mu)} + \gamma L_k^{(\sigma)}$ for a fixed $k$.

## A. Expression for $L_k^{(\mu)}$

In order to find a closed expression for the optimal readout layer $\theta_k$ after purification, we can first focus on the term $L_k^{(\mu)}$. We can write it as

$$L_k^{(\mu)} = \sum_i u_{i,k}^{(\mu)} \int_G \mathrm{d}\mu(q) \left\| \rho_k(q^{-1}) \hat{y}^{(k)}(qx_i) - y_i^{(k)} \right\|^2 = \sum_i u_{i,k}^{(\mu)} \int_G \mathrm{d}\mu(q) \left\| \theta_k^T \phi(qx_i) - \rho_k(q) y_i^{(k)} \right\|^2, \tag{84}$$

by using the invariance of the $\ell^2$ norm under the action of any orthogonal representation and by using the defining property of the representations. This expression is already standard and in the form of a least-squares regression problem in $\theta_k$. However, in order to put it on a form which is more suitable to be combined with the second term of the loss, we rewrite it in its equivalent vectorized form. This is done by defining $\vartheta_k := \text{vec}(\theta_k^T) \in \mathbb{R}^{d_k(d_{\text{model}}+1)}$ and by using the property of the vectorization operator for the product of three arbitrary matrices such that $\text{vec}(ABC) = (C^T \otimes A)\text{vec}(B)$, where $\otimes$ is the Kronecker product. In this way we obtain

$$\text{vec}(\theta_k^T \phi(qx_i)) = \text{vec}(\mathbb{I}_{d_k} \theta_k^T \phi(qx_i)) = \left(\phi(qx_i)^T \otimes \mathbb{I}_{d_k}\right) \vartheta_k. \tag{85}$$

Using the fact that $\|a\|^2 = \|\text{vec}(a)\|^2$ for any matrix $a$, we can write the loss as

$$L_k^{(\mu)} = \sum_i u_{i,k}^{(\mu)} \int_G \mathrm{d}\mu(q) \left\| \left(\phi(qx_i)^T \otimes \mathbb{I}_{d_k}\right) \vartheta_k - \rho_k(q) y_i^{(k)} \right\|^2 = \vartheta_k^T \Sigma_k^{(\mu)} \vartheta_k - 2\vartheta_k^T b_k^{(\mu)} + \text{const.}, \tag{86}$$

where we omitted all the terms that do not depend on $\vartheta_k$ as they are irrelevant for the optimization, and where we defined

$$\Sigma_k^{(\mu)} := \left[ \int_G \mathrm{d}\mu(q) \left( \sum_i u_{i,k}^{(\mu)} \phi(qx_i)\phi(qx_i)^T \right) \right] \otimes \mathbb{I}_{d_k}, \qquad \text{and} \qquad b_k^{(\mu)} := \sum_i u_{i,k}^{(\mu)} \int_G \mathrm{d}\mu(q) \left( \phi(qx_i) \otimes \mathbb{I}_{d_k} \right) \rho_k(q) y_i^{(k)}. \tag{87}$$

We recognize the quantity in the round brackets of $\Sigma_k^{(\mu)}$ as being the weighted uncentered second-moment matrix of the transformed $\phi(qx_i)$ over the samples $i$ for a fixed transformation $q$, lifted in the space of the parameters. Therefore we can interpret the term $L_k^{(\mu)}$ as expressed in terms of the group-averaged second-moment operator $\Sigma_k^{(\mu)}$ and of the corresponding cross-covariance vector $b_k^{(\mu)}$ between the featurized input and the target.

### B.  Expression for $L_k^{(\sigma)}$

For the second term of the loss, we essentially follow the same approach as for the first term. In this case, the vectorization procedure again exposes the quadratic dependence on $\vartheta_k$, which is crucial to find a closed form for the optimal readout layer after purification. In terms of $\theta_k$ and $\phi(x_i)$, the loss can be written as

$$L_k^{(\sigma)} = \sum_i u_{i,k}^{(\sigma)} \int_G \mathrm{d}\mu(q) \left\| \rho_k(q^{-1})\theta_k^T \phi(qx_i) - \int_G \mathrm{d}\mu(q')\, \rho_k(q'^{-1})\theta_k^T \phi(q'x_i) \right\|^2. \tag{88}$$

Using again the vectorization operator and the property $\mathrm{vec}(ABC) = (C^T \otimes A)\mathrm{vec}(B)$, we can write

$$\mathrm{vec}\big(\rho_k(q^{-1})\theta_k^T \phi(qx_i)\big) = \big(\phi(qx_i)^T \otimes \rho_k(q^{-1})\big)\,\vartheta_k, \tag{89}$$

which extracts the vectorized parameters $\vartheta_k$. Applying this identity to both terms of the loss, we can expose its quadratic nature and write it as

$$L_k^{(\sigma)} = \sum_i u_{i,k}^{(\sigma)} \int_G \mathrm{d}\mu(q) \left\| \left[ \big(\phi(qx_i)^T \otimes \rho_k(q^{-1})\big) - \int_G \mathrm{d}\mu(q')\, \big(\phi(q'x_i)^T \otimes \rho_k(q'^{-1})\big) \right] \vartheta_k \right\|^2. \tag{90}$$

which can be written as

$$L_k^{(\sigma)} = \vartheta_k^T S_k^{(\sigma)} \vartheta_k, \tag{91}$$

with

$$S_k^{(\sigma)} := \int_G \mathrm{d}\mu(q) \left[ \sum_i u_{i,k}^{(\sigma)} \big(\Phi_k(x_i,q) - \langle \Phi_k(x_i)\rangle_G\big)^T \big(\Phi_k(x_i,q) - \langle \Phi_k(x_i)\rangle_G\big) \right]. \tag{92}$$

where we introduced the "pulled-back" feature map $\Phi_k(x_i,q)$ and its average over the group $\langle \Phi_k(x_i)\rangle_G$, defined as

$$\Phi_k(x_i,q) := \phi(qx_i)^T \otimes \rho_k(q^{-1}), \qquad \text{and} \qquad \langle \Phi_k(x_i)\rangle_G := \int_G \mathrm{d}\mu(q)\, \Phi_k(x_i,q). \tag{93}$$

Again, we can recognize the quantity in square brackets in $S_k^{(\sigma)}$ as a weighted second-moment operator of the orbit-centered pulled-back feature map. Therefore, the term $L_k^{(\sigma)}$ can be interpreted in terms of the group-averaged centered second moment of the pulled-back feature map.

### C.  Closed form for the optimal readout

Putting together the two terms of the loss, we can write the total loss for the channel $k$ as

$$L_k = \vartheta_k^T \left( \Sigma_k^{(\mu)} + \gamma S_k^{(\sigma)} \right) \vartheta_k - 2\vartheta_k^T b_k^{(\mu)} + \text{const.}, \tag{94}$$

which is a quadratic function of $\vartheta_k$. Therefore, the optimal coefficients are obtained by the minimum of this quadratic function, which is given by the closed form expression

$$\vartheta_k^* = \left( \Sigma_k^{(\mu)} + \gamma S_k^{(\sigma)} \right)^{-1} b_k^{(\mu)}. \tag{95}$$

Reverting the vectorization process, we can then obtain the optimal readout layers $\theta_k^*$ after purification for each channel $k$. We remark that the evaluation of this expression requires only Eqs. (87) and (92), which can be interpreted as averages over second moments. As these involve an average over the group, they can be efficiently evaluated by using appropriate quadrature schemes such the one described in the previous section.

As a final remark we observe that the evalution of all these quantities are greatly simplified in the case of a 1-dimensional representation, as the Kronecker products are also done with 1-dimensional matrices (scalars), practically disappearing from the expressions.

## VII. PERTURBATIVE EXPANSION FOR THE LOGITS

This section shows that the pseudo-scalar channel is expressed as a "third order effect" at initialization in an architecture like the one of PET-MAD [7]. The model takes as input a point cloud with $N$ points, and for each point $i$ it takes as input the list of edges $\{r_{ji}\}$ and associated species (colors) $\{z_j\}$, where $r_{ji} = r_j - r_i$ is the relative position vector between the point $j$ and the point $i$. The model then applies a non-linear transformation to this list of edges and colors to produce a list of abstract tokens $\{t_{ji}\}$, with dimensionality $d_{\text{model}}$, which are then processed by a sequence of graph neural network (GNN) layers connected by message passing. Tokens are produced at each GNN layer, and each token $t_{ji}$ depends only on the edge $r_{ji}$, the color $z_j$ and the message received from the previous GNN layer. We now observe that in order to express a pseudo scalar ($\sigma = -1$) from vectors we need a dependence on at least two independent vectors. The requirement is even more strict for pseudoscalar channels ($\sigma = -1$, $\lambda = 0$), which require a dependence on at least three independent vectors [8]. Therefore we can already observe that the tokens for the initial layer are only a function of the edge and the color, and therefore they cannot express any pseudo ($\sigma = -1$) channel. Deeper in the architecture, each of the GNN layer contains an attention mechanism which connects all the edges of the same point $\{i\}$ in a non-linear way. Thus, this is the step in which the tokens can start to express all the pseudo channels. Here we will follow a simplified approach, where the tokens are first transformed into queries and values, $t_{ji} \xrightarrow{Q} q_{ji}$ and $t_{ji} \xrightarrow{V} v_{ji}$, and then the logit for the attention is computed as [9]

$$l_{ji} = \sum_k \frac{q_{ki}^T v_{ji}}{\sqrt{d_{\text{model}}}} + b_{ji}, \tag{96}$$

where the term $b_{ji}$ takes into account any masking or biases. We start by the empirical evidence that at initialization, the scalar channels are the most strongly expressed, with all the others being much smaller: we found that higher angular channels follow an exponential decay, as shown in Fig. 3 in the main text. We model this observation by assuming a perturbative expansion of the logit, as

$$l_{ji} = c_{ji} + \epsilon u_{ji}, \tag{97}$$

where $c_{ji}$ is the dominant scalar contribution and $u_{ji}$ contains all the remaining isotypic channels. To make the expansion clearer we introduce the small parameter $\epsilon$, which indicates the relative magnitude of the non-scalar channels with respect to the scalar one. We can now expand the softmax in the attention mechanism as

$$a_{ji} = \frac{\exp\{l_{ji}\}}{\sum_k^{\text{edges}} \exp\{l_{ki}\}} = \frac{\exp\{c_{ji} + \epsilon u_{ji}\}}{\sum_k^{\text{edges}} \exp\{c_{ki} + \epsilon u_{ki}\}} = a_{ji}^{(0)} \exp\{\epsilon u_{ji}\} \frac{1}{S(\epsilon)}, \tag{98}$$

where the last expression is defined in terms of the scalar attention $a_{ji}^{(0)} := \exp\{c_{ji}\} / \sum_k \exp\{c_{ki}\}$ and the normalization factor $S(\epsilon) := \sum_k a_{ki}^{(0)} \exp\{\epsilon u_{ki}\}$. We can also define the moments as $\mu_{i,n} := \sum_k a_{ki}^{(0)} u_{ki}^n$, and notice that $\mu_{i,0} = 1$. With this, using the Taylor expansion the function $1/(1+x)$ and of the exponential we can write

$$\frac{1}{S(\epsilon)} = \frac{1}{\sum_k a_{ki}^{(0)} \exp\{\epsilon u_{ki}\}} = \frac{1}{1 + \left(\sum_{n=1}^{\infty} \frac{\epsilon^n}{n!} \mu_{i,n}\right)} = 1 - \epsilon \mu_{i,1} + \epsilon^2 \left(\mu_{i,1}^2 - \frac{\mu_{i,2}}{2}\right) + \epsilon^3 \left(-\mu_{i,1}^3 + \mu_{i,1}\mu_{i,2} - \frac{\mu_{i,3}}{6}\right) + O(\epsilon^4). \tag{99}$$

Now, expanding also the remaining exponential we obtain

$$a_{ji} = a_{ji}^{(0)} + \epsilon a_{ji}^{(1)} + \epsilon^2 a_{ji}^{(2)} + \epsilon^3 a_{ji}^{(3)} + O(\epsilon^4), \tag{100}$$

where

$$a_{ji}^{(1)} := a_{ji}^{(0)}(u_{ji} - \mu_{i,1}), \tag{101}$$

$$a_{ji}^{(2)} := a_{ji}^{(0)} \left(\frac{u_{ji}^2}{2} - u_{ji}\mu_{i,1} + \mu_{i,1}^2 - \frac{\mu_{i,2}}{2}\right), \tag{102}$$

$$a_{ji}^{(3)} := a_{ji}^{(0)} \left(\frac{u_{ji}^3}{6} - \frac{u_{ji}^2}{2}\mu_{i,1} + u_{ji}\left(\mu_{i,1}^2 - \frac{\mu_{i,2}}{2}\right) - \mu_{i,1}^3 + \mu_{i,1}\mu_{i,2} - \frac{\mu_{i,3}}{6}\right). \tag{103}$$

| | |
|---|---|
| cutoff | 4.5 |
| cutoff_function | Bump |
| cutoff_width | 0.5 |
| d_pet | 128 |
| d_head | 128 |
| d_node | 256 |
| d_feedforward | 256 |
| num_heads | 8 |
| num_attention_layers | 2 |
| num_gnn_layers | 2 |
| normalization | RMSNorm |
| activation | SwiGLU |
| attention_temperature | 1.0 |
| transformer_type | PreLN |
| featurizer_type | feedforward |

TABLE I. Hyperparameters of the trained PET model.

Recalling that pseudo channels require a dependence on at least two independent vectors, and pseudoscalar channels require a dependence on at least three independent vectors, we can see that the first order $a_{ji}^{(1)}$ is able to express only proper channels. The second order $a_{ji}^{(2)}$ is able to express also pseudo channels, in the terms $\mu_{i,1}^2$ and $u_{ji}\mu_{i,1}$, but it cannot express pseudoscalar channels. Finally, the third order $a_{ji}^{(3)}$ is able to express also pseudoscalar channels, in the terms $\mu_{i,1}^3$ and $u_{ji}\mu_{i,1}^2$. Therefore we can conclude that the pseudo channels are expressed as a second order effect at initialization, while the pseudoscalar channels are expressed as a third order effect at initialization.

From the empirical fact that the trends analyzed here are preserved across the architecture, we can conclude that the pseudo channels are expected to be smaller than the proper ones, with a significant gap between them, while the pseudoscalar channels are expected to be even smaller, with a significant gap with respect to the pseudo channels, which is consistent with the results shown in Fig. 3 of the main text.

## VIII. THE POINT-EDGE TRANSFORMER ARCHITECTURE

A full description of the modules that comprise the latest version of the PET architecture is shown in Figure 1. For the MLIP results in the main text, PET was trained with `metatrain` version 2026.2.1 with the model hyperparameters given in Table I

## IX. POINT-WISE CHARACTER DECOMPOSITION FOR POLAR-MAE

In Figure 2 we report the point-wise normalized character decompositions for the same representative event shown in the main text. From the figure it is possible to appreciate the dominant scalar contribution and the quenching of the pseudotensorial space. As for the PET case, the pseudoscalar contribution are the most suppressed.

---

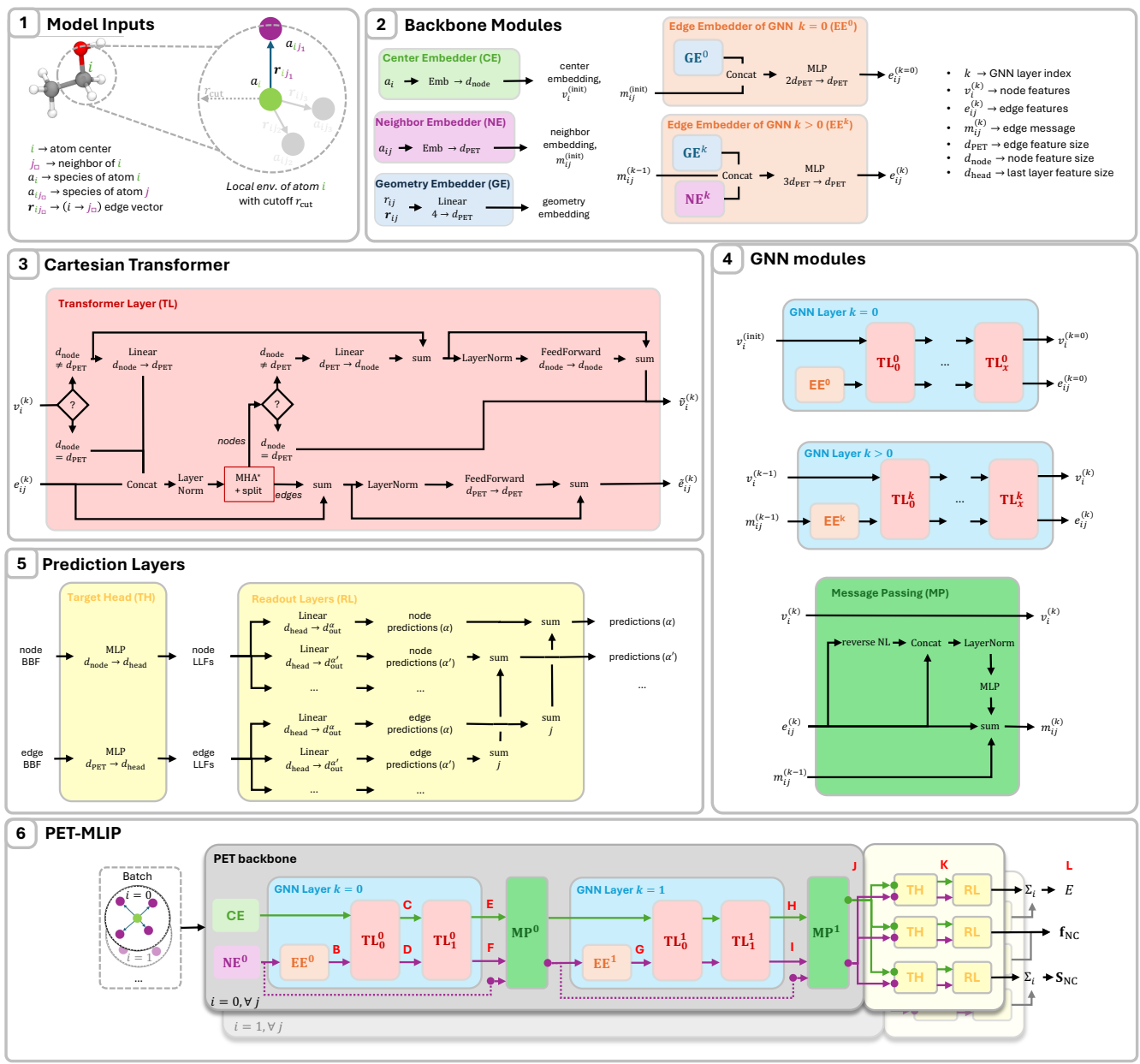

FIG. 1. Overview of the modules that comprise the PET architecture and the specific MLIP architecture used in this work.

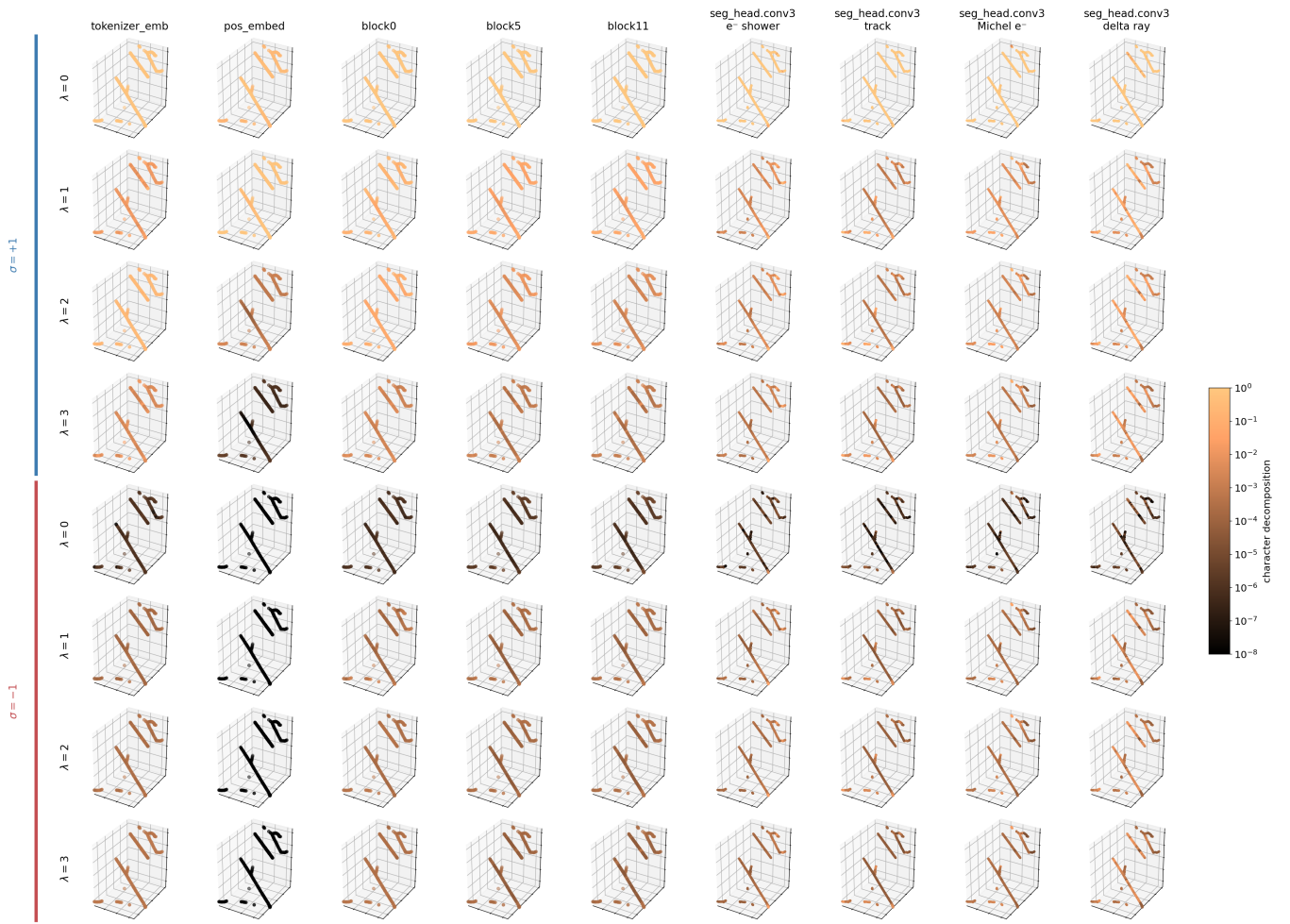

FIG. 2. Full point-wise character decompositions for a representative event.



\end{document}